\documentclass[letterpaper,11pt]{article}

\usepackage[utf8]{inputenc}

\usepackage[T1]{fontenc}
\usepackage{kpfonts}
\usepackage{microtype}

\usepackage[table]{xcolor}

\usepackage{arxiv}

\usepackage{url}            
\usepackage{booktabs}       
\usepackage{amsfonts}       
\usepackage{amssymb}
\usepackage{nicefrac}       
\usepackage{microtype}      
\usepackage{amsmath,amsthm}
\usepackage{color}
\usepackage{xspace}
\usepackage{wrapfig}
\usepackage{subfigure} 
\usepackage{capt-of}
\usepackage[list=off]{caption}
\usepackage{graphicx}

\usepackage{natbib}
\usepackage{algorithm}
\usepackage{algorithmic}
\usepackage{fancyhdr}

\usepackage{enumitem}

\usepackage[font=small,labelfont=bf,tableposition=top]{caption}
\DeclareCaptionLabelFormat{andtable}{#1~#2  \&  \tablename~\thetable}
\definecolor{DarkGreen}{rgb}{0.1,0.5,0.1}
\definecolor{DarkRed}{rgb}{0.5,0.1,0.1}
\definecolor{DarkBlue}{rgb}{0.1,0.1,0.5}

\usepackage{hyperref}       
\hypersetup{backref=true,       
    pagebackref=true,               
    hyperindex=true,                
    colorlinks=true,                
    breaklinks=true,                
    urlcolor= black,                
    linkcolor= DarkBlue,                
    bookmarks=true,                 
    bookmarksopen=false,
    filecolor=black,
    citecolor=DarkGreen,
    linkbordercolor=blue
}

\usepackage[font=normalsize,labelfont=bf, labelsep=colon, textfont=it]{caption}

\def\argdot{{\hspace{0.18em}\cdot\hspace{0.18em}}}

\newcommand{\cocoa}{\textsc{CoCoA}\xspace}

\DeclareMathOperator*{\argmin}{arg\,min}

\newcommand{\R}{\mathbb{R}}                      
\newcommand{\Exp}{\mathbb{E}}                      

\newcommand{\xv}{ {\bf x}}
\newcommand{\yv}{ {\bf y}}

\newcommand{\vv}{ {\bf v}}

\newcommand{\cv}{ {\bf c}}

\newcommand{\ev}{ {\bf e}}

\newcommand{\cP}{\mathcal{P}}

\newcommand{\xvt}{{\xv_{t}}}
\newcommand{\xvtp}{{\xv_{t+1}}}

\newcommand{\xvs}{{\xv^{\star}}}
\newcommand{\vvs}{{\vv^{\star}}}
\newcommand{\Dxv}{{\Delta \xv}}

\newcommand{\vvt}{{\vv_{t}}}

\newcommand{\cPt}{{\cP^{t}}}
\newcommand{\cPts}{\cP^{t}} 

\theoremstyle{plain}
\newtheorem{theorem}{Theorem}
\newtheorem{lemma}[theorem]{Lemma}

\newtheorem{assumption}{Assumption}

\theoremstyle{definition}
\newtheorem{definition}{Definition}

\newcommand{\vt}{{\bf v}_t}

\newcommand\greencell{\cellcolor{green!10}}
\newcommand\bluecell{\cellcolor{blue!10}}
\newcommand\redcell{\cellcolor{red!10}}
\newcommand\graycell{\cellcolor{black!10}}

\usepackage[affil-it]{authblk} 
\title{Randomized Block-Diagonal Preconditioning for Parallel Learning}

\author[1]{Celestine Mendler-D\"unner}
\author[2]{Aurelien Lucchi}
\affil[1]{University of California, Berkeley}
\affil[2]{ETH Z\"urich}

\date{}

\usepackage{hyperref}

\begin{document}

\maketitle

\vspace{1.5em}

\begin{abstract}
We study preconditioned gradient-based optimization methods where the preconditioning matrix has block-diagonal form. Such a structural constraint comes with the advantage that the update computation is block-separable and can be parallelized across multiple independent tasks. Our main contribution is to demonstrate that the convergence of these methods can significantly be improved by a randomization technique which corresponds to repartitioning coordinates across tasks during the optimization procedure. We provide a theoretical analysis that accurately characterizes the expected convergence gains of repartitioning and validate our findings empirically on various traditional machine learning tasks. From an implementation perspective, block-separable models are well suited for parallelization and, when shared memory is available, randomization can be implemented on top of existing methods very efficiently to improve convergence.
\end{abstract}


\section{Introduction}
\label{sec:intro}

We focus on the task of parallel learning where we want to solve the convex and smooth optimization problem 
\begin{equation}
\min_{\xv\in\R^n} f(\xv)
\label{eq:obj}
\end{equation}
on a multi-core machine with shared memory. In this context we study iterative optimization methods where the repeated computation of the incremental update 
\begin{equation}
\xvtp\leftarrow \xvt + \Dxv
\label{eq:update}
\end{equation}
is parallelized across cores. Such methods traditionally build on one of the following three approaches: i) they implement a {mini-batch algorithm} \citep{minibatchsgd}, where a finite sample approximation to $f$ is used to compute the update $\Dxv$, ii) they implement {asynchronous updates} \citep{Niu:2011wo,Liu:2015wj}, where stochastic updates are interleaved, or iii) they compute {block updates} \citep{Richtarik:2012vf}, where multiple coordinates of $\xv$ are updated independently and in parallel. The primary goal of all these methods is to introduce parallel computations into an otherwise stochastic algorithm in order to better utilize the number of available cores.

\citet{syscd} argue that these traditional methods are often not able to utilize the full potential of parallel systems because they make simplified modeling assumptions of the underlying hardware: They treat a multi-core machine as a uniform collection of cores whereas in fact it is a more elaborate system with complex data access patterns and cache structures. As a consequence, memory contention issues and false sharing can significantly impede their performance. 

To resolve this the authors have built on ideas from \emph{distributed} learning and proposed a novel approach to \emph{parallel} learning  that relies on a block-separable auxiliary model.
The imposed structure has the advantage that, in addition to computational parallelism, it implements a stricter separability between computational tasks which enables more efficient implementations. 
However, this does not come for free -- enforcing block-separability can slow down convergence as typically observed in distributed methods.

Interestingly, the new application area of block-separable models in a single machine setting with shared memory opens the door to previously unstudied algorithmic optimization techniques that can help counteract the convergence slow-down.  Namely, we can relax strong communication constraints, as long as we do not compromise the desired separability between computational tasks. 

One such algorithmic technique that preserves separability and can help convergence is \textit{repartitioning}. It refers to randomly assigning coordinates to tasks for each update step.
In a distributed setting repartitioning would involve expensive communication of large amounts of data across the network and has thus not been considered as an option. But in a single machine setting we can reassign coordinates to cores without incurring significant overheads.
This has been verified empirically by \citet{syscd} who showed that for the specific example of training a logistic regression classifier using the \cocoa method \citep{Smith:2016wp} the convergence gains of repartitioning can significantly prevail the overheads of reassign coordinates to cores in a shared-memory setting.

In this work we follow up on this interesting finding and provide the first theoretical study of repartitioning. In particular, we frame repartitioning as a randomization step applied to a block-diagonal preconditioning matrix. This allows us to quantify the gain of repartitioning over static partitioning in a general framework, covering a broad range of block-separable methods found in the distributed learning literature, including the \cocoa method. We further validate our theoretical findings empirically for both ridge regression and logistic regression on a variety of datasets with different sizes and dimensions. Both our theoretical and empirical results indicate that repartitioning can significantly improve the sample efficiency of a broad class of distributed algorithms and thereby turn them into interesting new candidates for parallel learning.

\section{Background}

We begin by providing some background on distributed learning methods.
This helps us set up a general framework for analyzing repartitioning in later sections.

\subsection{Distributed Optimization}
\label{sec:dist}

Distributed optimization methods are designed for the scenario where the training data is too large to fit into the memory of a single machine and thus needs to be stored in a distributed fashion across multiple nodes in a cluster.
The main objective when designing a distributed algorithm is to define an optimization procedure such that each node can compute its part of the update \eqref{eq:update} independently. In addition, this computation should only require access to local data and rely on minimal interaction with other workers. 

There are different approaches to achieve this computational separability. They all rely on a second-order approximation to the objective $f$ around the current iterate $\xvt$:
\begin{align}
f(\xvt+\Dxv) &\approx \tilde f_\xvt(\Dxv;Q_t)\label{eq:obj2}\\
&:=f(\xvt) + \nabla f(\xvt)^\top \Dxv + \frac 1 {2} \Delta \xv^\top Q_t \Delta\xv.\notag
\end{align}
Note that  $Q_t\in \R^{n\times n}$ can be a function of the iterate $\xvt$.
In a single machine case the optimal choice for $Q_t$ would be the Hessian matrix $\nabla^2 f(\xvt)$. But in a distributed setting we can not, or do not want to compute and store the full Hessian matrix across the entire dataset. 

One approach to nevertheless benefit from second-order information is to locally use a finite sample approximation to $\nabla^2 f(\xvt)$ for computing $\Dxv^k$  on each machine $k\in[K]$, before aggregating these updates to get a global update $\Dxv$. This strategy has been exploited in methods such as DANE \citep{shamir2014communication}, GIANT \citep{wang2017giant}, AIDE \citep{reddi2016aide} and DISCO \citep{zhang2015disco}. The convergence of these methods typically relies on concentration results which require the data to be distributed uniformly across the machines. Otherwise, their analysis is indifferent to the specifics of the data partitioning.

For studying repartitioning we focus on an orthogonal approach, where the computation of the individual coordinates of $\Dxv$ is distributed across machines. This includes methods such as \cocoa \citep{Smith:2016wp}, ADN \citep{duenner2018adn} and other distributed block coordinate descent methods such as \citep{lee2017distributed,Hsieh:2016wg,mahajan2017distributed,lee2017distributed}. All these methods construct a separable auxiliary model of the objective function by enforcing a block-diagonal structure on $Q_t$ in \eqref{eq:obj2}. As illustrated in Figure~\ref{fig:separability}, this renders the computation of the individual coordinate blocks of $\Dxv$ independent.

\begin{figure*}[t!]
\begin{equation*}
\setlength\arraycolsep{4pt}
\small
\begin{array}{c}
\underbrace{\left[\begin{array}{>{\columncolor{green!10}}c}
\cdot\\[2pt]
\cdot\\[2pt]
\cdot\\[2pt]
\cdot\\[2pt]
\cdot\\[2pt]
\cdot
\end{array}\right]}_{\textstyle\xvtp}
=
\underbrace{\left[\begin{array}{>{\columncolor{green!10}}c}
\cdot\\[2pt]
\cdot\\[2pt]
\cdot\\[2pt]
\cdot\\[2pt]
\cdot\\[2pt]
\cdot
\end{array}\right]}_{\textstyle\xvt}-\eta\;
{\underbrace{
\left[\begin{array}{cccccc}
\bluecell\cdot&\bluecell\cdot&0&0&0&0\\[2pt]
\bluecell\cdot&\bluecell\cdot&0&0&0&0\\[2pt]
0&0&\redcell\cdot&\redcell\cdot&0&0\\[2pt]
0&0&\redcell\cdot&\redcell\cdot&0&0\\[2pt]
0&0&0&0&\graycell\cdot&\graycell\cdot\\[2pt]
0&0&0&0&\graycell\cdot&\graycell\cdot\\
\end{array}\right]}_{\textstyle Q_\cPt}}^{-1}
\underbrace{\left[\begin{array}{>{\columncolor{green!10}}c}
\cdot\\[2pt]
\cdot\\[2pt]
\cdot\\[2pt]
\cdot\\[2pt]
\cdot\\[2pt]
\cdot
\end{array}\right]}_{\textstyle\nabla f(\xvt)}
\end{array}
\quad\Rightarrow\quad
\begin{array}{ccccccc}
\left[\begin{array}{>{\columncolor{green!10}}c}
\cdot\\
\cdot
\end{array}\right] &=&\left[\begin{array}{>{\columncolor{green!10}}c}
\cdot\\
\cdot
\end{array}\right] &-&\eta&\left[\begin{array}{>{\columncolor{blue!10}}c >{\columncolor{blue!10}}c}
\cdot&\cdot\\
\cdot&\cdot
\end{array}\right]^{-1}&\left[\begin{array}{>{\columncolor{green!10}}c}
\cdot\\
\cdot
\end{array}\right] \\[10pt]
\left[\begin{array}{>{\columncolor{green!10}}c}
\cdot\\
\cdot
\end{array}\right] &=&\left[\begin{array}{>{\columncolor{green!10}}c}
\cdot\\
\cdot
\end{array}\right] &-&\eta&\left[\begin{array}{>{\columncolor{red!10}}c >{\columncolor{red!10}}c}
\cdot&\cdot\\
\cdot&\cdot
\end{array}\right]^{-1}&\left[\begin{array}{>{\columncolor{green!10}}c}
\cdot\\
\cdot
\end{array}\right] \\[10pt]
\underbrace{\left[\begin{array}{>{\columncolor{green!10}}c}
\cdot\\
\cdot
\end{array}\right]}_{\textstyle\xvtp} &=&\underbrace{\left[\begin{array}{>{\columncolor{green!10}}c}
\cdot\\
\cdot
\end{array}\right]}_{\textstyle\xvt} &-&\eta&{\underbrace{\left[\begin{array}{>{\columncolor{black!10}}c >{\columncolor{black!10}}c}
\cdot&\cdot\\
\cdot&\cdot
\end{array}\right]}_{\textstyle{Q_t}_{[\cPts_k,\cPts_k]}}}^{-1}&\underbrace{\left[\begin{array}{>{\columncolor{green!10}}c}
\cdot\\
\cdot
\end{array}\right] }_{\textstyle\nabla f(\xvt)}
\end{array}
\end{equation*}
\vspace{-0.2cm}
\caption{Parallelism in the update computation induced by block-diagonal pre-conditioning as decribed in Algorithm~\ref{alg:algo}.}
\label{fig:separability}
\end{figure*}

The partitioning of the coordinates across nodes determines which off-diagonal elements of $Q_t$ are being ignored. 
While not all elements of $Q_t$ might be equally important, each partitioning inevitably ignores a large subset of elements which can hurt convergence.
Repartitioning offers an interesting alternative. It considers a different subset of elements for each update step and over the course of the algorithm it gets information from all elements of $Q_t$ with non-zero probability. 

To gain intuition how repartitioning helps convergence, let us investigate the specific structure of the matrix $Q_t$ at the example of generalized linear models (GLMs).

\subsubsection{GLM Training}
\label{sec:glms}

One attract of GLMs is the simple linear dependence of the objective function on the data matrix imposed by the model. This makes GLMs particularly appealing in distributed settings where tasks can be separated across data partitions. It most likely also explains why so many distributed methods found in the literature have been specifically designed for GLMs. 

For GLMs, the objective $f$ depends linearly on the data matrix $A\in\R^{m\times n}$:
\begin{equation}
f(\xv)=\ell(A \xv),
\label{eq:glm}
\end{equation}  
where $\ell:\R^m\rightarrow \R$ in general denotes the loss function.
Let $\vv_t := A\xv_t$ be the information that is periodically shared across nodes, then the second-order model $\tilde f_\xvt(\Dxv;Q_t)$ in \eqref{eq:obj2} can be written as
\begin{equation}
\ell(\vvt) + \nabla \ell(\vvt)^\top A \Dxv  + \frac 1 {2} \Delta \xv^\top Q_t  \Delta\xv \quad \text{where}\quad Q_t=A^\top \hat Q_t A.
\label{eq:objglm}
\end{equation}
with the optimal choice of $\hat Q_t = \nabla^2\ell(\vt)$. An appealing property of \eqref{eq:objglm} is that the GLM structure makes the dependence of $\tilde f_\xvt$ (and hence $Q_t$) on the data more explicit. It is not hard to see that separability of \eqref{eq:objglm} across coordinate blocks and corresponding columns of $A$ follows by making $Q_t$ block-diagonal and setting elements outside the diagonal blocks to zero.  This is particularly easy to realize if $\hat Q_t$ has diagonal form.
This observation has been the basis for many distributed methods such as \cocoa \citep{Smith:2016wp,Jaggi:2014vi}, {ADN} \citep{duenner2018adn} and other block separable methods such as \citep{lee2017distributed}. In \cocoa, $\hat Q_t$ is set to $\gamma_\ell I$ -- where $\gamma_\ell$ denotes the smoothness parameter of $\ell$ -- thus forming an over-approximation to $\nabla^2 \ell$. In \citep{duenner2018adn} and \citep{lee2017distributed}, it was observed that for popular loss functions used in machine learning, $\nabla^2 \ell(\xvt)$ is a diagonal matrix. Hence, they keep $\hat Q_t = \nabla^2 \ell(\xvt)$ and directly enforce the block-diagonal structure on $A^\top \nabla^2\ell(\xvt)A$ to preserve additional local second-order information. We note that methods of the latter form are augmented by a line-search strategy or a trust-region like approach~\citep{nesterov2006cubic} to guarantee sufficient function decrease and ensure convergence. We will come back to this condition in Section \ref{sec:general}.

\subsubsection{Static Partitioning}
\label{sec:static}

Traditionally, distributed methods assume a static partitioning of data across nodes. In that way, expensive communication of data across the network can be avoided.
When distributing the computation of $\Dxv$ coordinate-wise across nodes, this implies a static allocation of data columns to nodes (hence coordinates to blocks) throughout the optimization. 

In this work, motivated by the recent trend in parallel learning, we take a different approach. We study the setting where one can randomly reassign coordinates to blocks for each repeated computation of $\Dxv$.  To the best of our knowledge, a formal study of such a repartitioning approach in the context of block separable methods does not yet exist in the literature.  This is likely due to the fact that block diagonal approximations to $Q_t$ have only been studied in the context of distributed learning where data repartitioning seems unreasonable.
In addition, the theoretical analysis of existing methods can not readily be extended to explain the effect of repartitioning because they look at the function decrease in each individual iteration in isolation. Therefore, we will resort to analysis tools from the literature on preconditioned gradient descent methods.

\subsection{Preconditioned Gradient Methods} 
Any optimization method that relies on a second-order auxiliary model of the form \eqref{eq:obj2} can be interpreted as a preconditioned gradient descent method where 
\begin{equation}
\xvtp = \xvt -  \eta Q_t^{-1} \nabla f(\xvt)
\label{eq:update2}
\end{equation}
and $\eta>0$ denotes the step size. Various choices for the matrix $Q_t$ have been discussed in the literature on preconditioning~\citep{convex_optimization_nocedal}. The simplest example is  standard gradient descent where $Q_t$ is equal to the identity matrix and $\eta$ is chosen to be inversely proportional to the smoothness parameter of $f$. On the other side of the spectrum, the classical Newton method defines $Q_t$ via the Hessian matrix $\nabla^2 f(\xvt)$. Since the computation of the exact Hessian is often too computationally expensive, even in a single machine setting, various approximation methods have been developed. Such methods typically rely on finite sample approximations to the Hessian using sketching techniques \citep{pilanci2016iterative}, sub-sampling~\citep{erdogdu2015convergence} or some quasi-Newton approximation \citep{dennis1977quasi}. They can also be combined with various line-search or trust-region-like strategies as in~\citep{blanchet2016convergence, kohler2017sub}. However, all these methods are either first-order methods, or they do not induce a block-diagonal structure on $Q_t$. Hence, our approach has also not been studied in the context of preconditioning methods until now.


\section{Method}

We introduce a general framework for studying repartitioning with the goal to cover the different distributed methods introduced in Section~\ref{sec:dist}. The common starting point in all these methods is a second-order approximations $\tilde f_\xvt(\argdot;Q_t)$ to $f$ as defined in \eqref{eq:obj2}. The methods then differ in their choice of $Q_t$ and the mechanisms they implement to guarantee sufficient function decrease when preconditioning on a block-diagonal version of $Q_t$. To focus on repartitioning in isolation we abstract these technicalities into assumptions  in Section~\ref{sec:conv}. For now, let us assume a \emph{good}  local second-order model  $\tilde f_\xvt(\argdot;Q_t)$  is given  and walk through the  block-diagonal preconditioning method outlined in Algorithm~\ref{alg:algo}. We first need to introduce some notation.

\subsection{Notation}
We write $i\in[n]$ for $i=1,2,...,n$ and we denote $\xvs=\argmin_\xv f(\xv)$ to refer to the minimizer of $f$ which is written as  $f^\star=f(\xvs)$.

\begin{definition}[Partitioning]\textit{
We denote the \emph{partitioning} of the indices $[n]$ into $K$ disjoint subsets as $\cP:=\{\cP_i\}_{i\in[K]}$  where $\cup_{i\in[K ]} \cP_i = [n]$ and $\cP_i \cap\cP_j = \emptyset, \; \forall i\neq j$.  If the partitioning is randomized throughout the algorithm we use the superscripts $\cPt$ to refer to the partitioning at iteration $t$.}
\end{definition}
Further, we write $\xv_{[\cP_k]}\in\R^n$ to refer to the vector with elements $(x_{[\cP_k]})_i=x_i$ for $i\in\cP_k$ and zero otherwise. Similarly, we use  $M_{[\cP_i,\cP_j]}\in\R^{n\times n}$ to denote the masked version of the matrix $M\in \R^{n\times  n}$, with only non-zero elements  for $M_{k,\ell}$ with $k\in \cP_i$ and $\ell\in\cP_j$.

\subsection{Block-Diagonal Preconditioning}
In each step $t\geq 0$ of Algorithm~\ref{alg:algo} we select a partitioning $\cPt$  and construct a block diagonal version $Q_\cPt$ from $Q_t$ according to $\cPt$:
\begin{equation}
Q_{\cPt} := \sum_{k\in[K]} {Q_t}_{[\cPts_k,\cPts_k]}.
\label{eq:separableQ}
\end{equation}
This matrix then serves as a preconditioning matrix in the update step (line 7) of Algorithm~\ref{alg:algo}. 
Note that  we will for illustration purposes, and without loss of generality, refer to $Q_\cPt$ as a block-diagonal matrix. 
Although $Q_\cPt$ is not necessarily block-diagonal for all $\cPt$, it can be brought into block-diagonal form by  permuting the rows and columns of the matrix.

\subsection{Dynamic Partitioning}

In a classical distributed method, the partitioning $\cPt$ is fixed throughout the entire algorithm, as discussed in Section~\ref{sec:static}. This corresponds to option (i) in Algorithm~\ref{alg:algo}. The novel feature in our study is to allow for a different random partitioning in each iteration $t$ and use the induced block diagonal preconditioning matrix to perform the update step.
This randomized procedure, also referred to as \textit{repartitioning}, is summarized as option (ii) in Algorithm~\ref{alg:algo}.


\begin{algorithm}[t!]
   \caption{\textit{Block-Diagonal Preconditioning} for \eqref{eq:obj}  with (i) static and (ii) dynamic partitioning}
   \label{alg:algo}
\begin{algorithmic}[1]
   \STATE {\bfseries Input:} $\tilde f_\xvt(\cdot,Q_t)$, step size $\eta$, partitioning $\cP$
\STATE \textbf{Initialize:} $\xv_0 \in \R^n$\vspace{0.1cm}
   \FOR{$t=0$ {\bfseries to} $T-1$}\vspace{0.1cm}
\STATE (i) use default partitioning $\cPt = \cP$
   \STATE (ii) choose a random partitioning $\cPt$ of the $i\in [n]$
\vspace{0.1cm}
	\FOR{$k\in{[K]}$ on each processor in parallel}
   \STATE $\xvtp \leftarrow \xvt - \eta Q_{[\cPts_k,\cPts_k]}^{-1}\nabla f(\xvt)$ 
   \ENDFOR
\vspace{0.1cm}
   \ENDFOR
\STATE \textbf{Return:} $\xv_T$
\end{algorithmic}
\end{algorithm}

\section{Convergence Analysis}
\label{sec:conv}

We now turn to the main contribution of our work which consists in analyzing and contrasting the convergence rate of Algorithm~\ref{alg:algo} for the two different partitioning techniques.
To convey our main message in the most transparent way, we start by analyzing a quadratic function where the second-order model  $\tilde f_\xvt$ in  \eqref{eq:obj2} is exact and no additional assumptions are needed. We then  extend this result to GLMs and to more general second-order auxiliary models. All proofs can be found in the appendix.


\subsection{Quadratic Functions}
Let us consider the setting where $f: \R^n \to \R$ is a quadratic function of the following form:
\begin{equation}
f(\xv) = \frac{1}{2} \xv^{\top} H \xv - \cv^{\top} \xv,
\label{eq:fquad}
\end{equation}
where $H\in \R^{n\times n}$ is a symmetric matrix and $\cv \in \R^n$.  The natural choice is to define $Q_t=H$ when working with the auxiliary model \eqref{eq:obj2}. Obviously, for quadratic functions using the full matrix $H$ for preconditioning would yield convergence in a single step. But under the constraint of Algorithm~\ref{alg:algo} that the preconditioning matrix $Q_\cPt$ has block diagonal structure this in general does not hold true. The convergence of Algorithm~\ref{alg:algo} for this setting  is explained in the following theorem:
\begin{theorem}
\label{thm:quad}
Assume $f$ is defined as \eqref{eq:fquad}, and $Q_t:=H$. Then Algorithm~\ref{alg:algo} with a fixed step size $\eta=\frac 1 K$  converges at a linear rate 
\begin{align*}
\Exp\left[f(\xvtp)-f(\xvs)\right] &\leq  \left(1-\rho\right)^t\left[f(\xv_0)-f(\xvs)\right]
\end{align*}
with
\begin{equation}
\rho:=\frac 1 K \lambda_{\min}\left(\Exp[H_\cPt^{-1}] H\right).
\label{eq:rho}
\end{equation}
The expectations are taken over the randomness of the partitioning $\cPt$.
\end{theorem}
One of the key insights is that \emph{the convergence rate of Algorithm~\ref{alg:algo} depends on the partitioning scheme through the term $\Exp[H_\cPt^{-1}]$}. Ideally, for optimal convergence we want $\rho$ to be as large as possible, and hence $\Exp[H_\cPt^{-1}]\approx H^{-1}$. How well $\Exp[H_\cPt^{-1}]$ is able to approximate $H^{-1}$ is measured by the spectrum of $\Exp[H_\cPt^{-1}] H$ which determines the convergence rate of the respective partitioning scheme. 
For fixed partitioning the expectation $\Exp[H_\cPt^{-1}]$ reduces to $H_\cP^{-1}$ induced by the default partitioning $\cP$, whereas for repartitioning it is an average over all possible partitionings. 
As a consequence the convergence of repartitioning is superior to static partitioning whenever the average of $H_\cP^{-1}$ over all partitionings is able to better approximates $H^{-1}$ compared to any individual term.

Note that even in cases where there exists a single fixed partitioning that is better than repartitioning, it could still be combinatorially hard to discover it and repartitioning provides an appealing alternative. We will provide several empirical results that support this claim in Section~\ref{sec:experiments} and  investigate the properties of the matrix $\Exp[H_\cPt^{-1}] H$ analytically for some particular $H$ in Section~\ref{sec:eff}.


\subsection{Smoothness Upperbound for GLMs}
\label{sec:glm}
As a second case study we focus on GLMs, as defined in \eqref{eq:glm}, where $\ell$ is a $\gamma_\ell$-smooth loss function. In this setting we analyze Algorithm~\ref{alg:algo} for the second-order model  $\tilde f_\xvt(\argdot;Q_t)$ defined through 
\begin{equation}
Q_t:= \gamma_\ell A^\top A.
\label{eq:Qglm}
\end{equation}
This model forms a global upper-bound on the objective function $f$. It is used in \citep{Smith:2016wp} and related algorithms. 
It can intuitively be understood that in this case the quality of a block-diagonal approximation $Q_{\cPt}$ depends on the correlation between data columns residing in different partitions; these are the coordinates of $Q_t$ that are being ignored when enforcing a block-diagonal structure for achieving separability.  Let us for simplicity denote $M:=A^\top A$. Then, the expected function decrease in each iteration of Algorithm~\ref{alg:algo} can be bounded as:
\begin{lemma}
\label{lem:fdecglm}
Assume $f$  has the form \eqref{eq:glm}, $ \ell$ is $\gamma_\ell$-smooth, and $Q_t$ is chosen as in \eqref{eq:Qglm}. Then, in each step $t\geq 0$ of Algorithm~\ref{alg:algo} with a fixed step size $\eta=\frac 1 {K\gamma_\ell}$ the objective decreases as 
\begin{equation*}
\Exp\left[f(\xvt)-f(\xvtp)\right] \geq  \frac 1 {2K\gamma_\ell}  \lambda_{\min}(A \Exp[M_\cPt^{-1}] A^{\top} ) \|\nabla \ell(A\xvt)\|^2
\end{equation*}
where expectation are taken over the randomness of the partitioning $\cPt$.
\end{lemma}

To translate Lemma~\ref{lem:fdecglm} into a convergence rate for Algorithm~\ref{alg:algo} we need a lower bound on the curvature of $f$ in order to relate the gradient norm to the suboptimality. The following standard assumption \citep{polyak} on the loss function $\ell$
allows us to do this:
\begin{assumption}[Polyak Lojasiewicz] 
\label{ass:PL}
Assume that for some $\mu_\ell>0$ the function $\ell$  satisfies   
\begin{equation}
\frac 1 2 \|\nabla \ell(\vv)\|^2\geq \mu_\ell (\ell(\vv)-\ell(\vvs)).
\label{eq:PL}
\end{equation}
where $\vvs:=\argmin_\vv \ell(\vv)$.
\end{assumption}
Note that this assumption is weaker than strong-convexity as shown by \citet{karimi2016linear}. The following linear convergence rate for  Algorithm~\ref{alg:algo} follows:
\begin{theorem}
\label{thm:glm}
Consider the same setup as in Lemma~\ref{lem:fdecglm} where $\ell$ in addition satisfies Assumption \ref{ass:PL} with constant $\mu_\ell>0$. 
Then, Algorithm~\ref{alg:algo} with a fixed step size $\eta=\frac 1 {K\gamma_\ell}$  converges as 
\begin{align*}
\Exp\left[f(\xvtp)-f(\xvs)\right] &\leq  \left(1-\rho\right)^t\left[f(\xv_0)-f(\xvs)\right].
\end{align*}
with
\begin{equation}
\rho:=\frac {\mu_\ell} {K\gamma_\ell} \lambda_{\min}\left(A \Exp[M_\cPt^{-1}] A^\top\right),
\label{eq:rho2}
\end{equation}
where $M_\cPt$ denotes the masked version of  $M:=A^\top A$  given by the partitioning $\cPt$. Expectations are taken over the randomness of the partitioning.
\end{theorem}

The dependence of the convergence rate in Theorem~\ref{thm:glm} on the partitioning is captured by the term $\lambda_{\min}(A \Exp[M_\cPt^{-1}]A^\top)$ which simplifies to $\lambda_{\min}( \Exp[M_\cPt^{-1}]M)$ for symmetric $A$. Similar to Theorem~\ref{thm:quad} this term reflects the penalty for using an inexact approximation to $M^{-1}$ in Algorithm~\ref{alg:algo}. The approximation quality is measured in expectation over all potential partitionings and is optimized (equal to $1$) for $ \Exp[M_\cPt^{-1}]= M^{-1}$.

Let us further comment on the dependence of  $\rho$ in \eqref{eq:rho2} on the condition number. In classical gradient descent, e.g., \citep{karimi2016linear}, the convergence rate depends linearly on the condition number of the objective function $f$, while our rate depends on $\kappa_\ell:={\mu_\ell}/{\gamma_\ell}$, the condition number of the loss function $\ell$. The smoothness and PL parameters of $f$ and $\ell$ relate as $\gamma_f =\gamma_\ell \lambda_{\max}\left(M\right)$ and $\mu_f =\mu_\ell \lambda_{\min}\left(M\right)$. 
Hence, in Theorem~\ref{thm:glm} the dependence on the condition number is improved by a factor of $ {\lambda_{\max}(M)}/{\lambda_{\min}(M)}\geq 1$ in comparison to gradient descent. The remaining factor of $\frac 1 K$ in front of the condition number is the price for using a conservative fixed step size in our analysis.


\subsection{General Auxiliary  Model}
\label{sec:general}
For the most general case we do not pose any structural assumption on $f$. We only assume the auxiliary model $\tilde f_\xvt(\argdot,Q_t)$ is a reasonably good approximation to the function $f$:
\begin{assumption}\label{ass:ub} $ \tilde f_\xvt (\argdot;Q_t)$ is such that $\forall \Dxv$ and some $\xi\in (0,1]$ it holds that
\begin{equation}
 f (\xvt +\Dxv )\leq \xi \tilde f_\xvt (\Dxv;Q_t)  + (1-\xi) f(\xvt).
\label{eq:ub}
\end{equation} 
\end{assumption}
Approximations $ \tilde f_\xvt (\argdot;Q_t)$ that satisfy \eqref{eq:ub} can be obtained for smooth functions by taking a Taylor approximation truncated at order $p$ and combined with a bound on the $p$-th derivative, see e.g.~\citep{birgin2017worst, nesterov2006cubic}. In our case where we want a quadratic model we choose $p=2$ and, because smoothness gives us an upper-bound, the inequality \eqref{eq:ub} holds for $\xi=1$.  Another popular approach to guarantee sufficient function decrease in the spirit of \eqref{eq:ub} are backtracking line-search methods, such as used in \citep{lee2017distributed}. Here $\xi$ directly maps to the control parameter in the Armijo-Goldstein condition \citep{armijo1966} if $Q_t$ is PSD.
In the appendix we elaborate on these connections and explain how our setting could be extended to also cover trust region like approaches \citep{Cartis2011} such as used in ADN \citep{duenner2018adn}.

For methods that build on auxiliary models that satisfy \eqref{eq:ub} we can quantify the dependence of the function decrease on the partitioning scheme using the following lemma.

\begin{lemma} 
\label{lem:general}
Consider a quadratic approximation $\tilde f_\xvt(\argdot;Q_t)$  satisfying Assumption~\ref{ass:ub}. Then, in each step of Algorithm~\ref{alg:algo} the function value decreases as
\begin{align*}
\Exp[ f(\xvt)-f(\xvtp)]
&\geq  \rho_t   \|{\Delta\tilde \xv_t}^\star\|^2
\end{align*}
where 
\begin{equation}
\rho_t:= \frac \xi {2K}  \lambda_{\min}(Q_t^\top \Exp[Q_{\cPt}^{-1}] Q_t ),
\label{eq:rho3}
\end{equation}
with ${\Delta\tilde \xv_t}^\star:= \argmin_\xv \tilde f_\xvt(\xv,Q_t)$ denoting the optimal next iterate according to $\tilde f_\xvt$.
\end{lemma}
Hence, even in the most general case, the dependency of the convergence rate on the partitioning scheme can be explained through a simple quantity involving the expected block-diagonal preconditioning matrix $\Exp[ Q_\cPt^{-1}]$: $\lambda_{\min}(Q_t^\top\Exp[ Q_\cPt^{-1}] Q_t )$. 

The auxiliary model $\tilde f_\xvt$, on the other hand, and hence its minimizer $\tilde \xvt^{\star}$ are independent of the partitioning. 
Hence, how we translate Lemma~\ref{lem:general} into a convergence results solely depends on the distributed method (the choice of $Q_t$) we deploy.
We make the following assumption:

\begin{assumption}[Sufficient function decrease]
\label{ass:function-dec}
The method defines $Q_t$ such that sufficient function decrease of the optimal  update ${\Delta\tilde \xv_t}^\star$ can be guaranteed:
\begin{equation}
f(\xvt + \Delta\tilde \xv_t^\star) - f(\xv^\star) \leq \alpha [f(\xvt) - f(\xv^\star)]
\label{ass:alpha}
\end{equation}
for some $\alpha\in [0,1)$. 
\end{assumption}

This assumption can be satisfied with a preconditioned gradient descent step and appropriate rescaling of $Q_t$ for any PSD matrix $Q_t$. Importantly, such a rescaling affects every partitioning scheme equally.

We note that alternative assumptions would also lead to convergence results. For example, techniques found in the trust-region and cubic regularization literature (see e.g.~\citep{Cartis2011, duenner2018adn})  have proposed to adapt the optimization algorithm instead to guarantee sufficient function decrease in the spirit of~\eqref{ass:alpha}.

Building on Assumption~\ref{ass:function-dec} we get  the following  rate of convergence for Algorithm~\ref{alg:algo}.

\begin{theorem}
\label{thm:general}
Assume $f$ is convex and  $L$-Lipschitz continuous, and the auxiliary model $ \tilde f_\xvt (\argdot;Q_t)$ satisfies Assumption~\ref{ass:ub} and Assumption~\ref{ass:function-dec}. Then, Algorithm~\ref{alg:algo} with a fixed step size $\eta= \frac{1}{K}$ converges as 
\begin{align*}
\Exp[f(\xvtp)-f(\xvs)] &\leq \left( 1- \min_t \rho_t \tfrac{(1-\alpha)}{L}  \right)^t \varepsilon_0
\end{align*}
where $\varepsilon_0 = f(\xv_0)-f(\xvs)$ and $\rho_t $ defined in \eqref{eq:rho3}.
\end{theorem}

Note that the step size $\eta = \frac 1 K $ is required throughout our analysis because we pose Assumption~\ref{ass:ub} on $\tilde f_\xv$ and need to guarantee convergence uniformly across partitionings for a method that uses a block diagonal version of $Q_t$. To dynamically adapt to each partitioning Algorithm~\ref{alg:algo} could be augmented with a line-search procedure. We omitted this to preserve clarity of our presentation. 

Similarly, all our results from this section can readily be extended to the case where the local subproblem \eqref{eq:obj2} is not necessarily solved exactly but only $\theta$-approximately (in the sense of  Assumption 1 used by  \citet{Smith:2016wp}). This provides additional freedom to trade-off overheads of repartitioning and sample efficiency for optimal performance.

\section{ Effect of Randomization}
\label{sec:eff}

Let us return to the quadratic case where the auxiliary model $\tilde f_\xvt$  is exact and focus on the dependence of  $\rho$ on the partitioning scheme. We recall that the value of $\rho$ as defined  in \eqref{eq:rho} is determined by the smallest eigenvalue of 
\begin{equation}
\Lambda_ \cP := Q_\cP^{-1} Q.
\label{eq:M}
\end{equation} 
In the following we will evaluate $\lambda_{\min}(\Exp[\Lambda_\cP])$ analytically for some particular choices of $Q$  to quantify the gain of repartitioning over static partitioning predicted by Theorem~\ref{thm:quad}. 

For simplicity, we assume $Q$ does not depend on $t$ and the partitioning $\cP$ is uniform, such that $|\cP_i|=|\cP_j| =n_k\,\forall j, i\in [K]$ and $n_k=\frac n K$ denotes the number of coordinates assigned to each partition.

\subsection{ Uniform Correlations}
\label{sec:uniform}
We start with the special case where all off-diagonal elements of $Q$ are equal to $\alpha\in[0,1)$. Thus, for $n=4$ the matrix $Q$ would look as follows: 
\begin{equation*}
\setlength\arraycolsep{3pt}
Q=\left[\begin{array}{cccccc}
1&\alpha&\alpha&\alpha\\
\alpha&1&\alpha&\alpha\\
\alpha&\alpha&1&\alpha\\
\alpha&\alpha&\alpha&1
\end{array}\right].
\end{equation*}
Such a structure of $Q$ would, for example, appear in a linear regression problem, where all columns of the data matrix $A$ are equally correlated. In such a scenario it does not matter which elements of $Q$ we ignore and all fixed partitionings are equivalent from an algorithmic perspective. We refer the reader to Figure~\ref{fig:uniform2} in the appendix  for an illustration of all matrices involved in this example.
Let us note that 
\[\Lambda_ \cP = Q_\cP^{-1}Q =Q_\cP^{-1}(Q_\cP+Q_\cP^c)= I + Q_\cP^{-1}Q_\cP^c \]
where $Q_\cP^c := Q - Q_\cP$. Given the specific structure of $Q$ considered in this example, the inverse $Q_\cP^{-1}$ can be computed from the individual blocks of $Q_\cP$ and is again symmetric and block-diagonal. As a consequence the diagonal blocks of $ Q_\cP^{-1}Q_\cP^c$ are zero and by symmetry all other elements are equal. We denote the value of these elements by $\epsilon$, where an exact derivation as a function of $\alpha$ can be found in Appendix~\ref{app:uniform}. 
In the following we will  evaluate 
\[\lambda_{\min}(\Exp[\Lambda_\cP]) = 1+ \lambda_{\min}(\Exp[Q_\cP^{-1}Q_\cP^c]) \]
for the case of static as well as dynamic partitioning.

\textit{(i) Static Partitioning.} We have $\Exp[\Lambda_\cP] = \Lambda_\cP$ and we compute $\lambda_{\min}(\Lambda_\cP)$ by exploiting the symmetry of the matrix $Q_\cP^{-1}Q_\cP^c $. The eigenvector corresponding to the smallest eigenvalue will be $\vv=\ev_{\cP_i}-\ev_{\cP_j}$ for any $i\neq j$ and the corresponding eigenvalue with multiplicity $K-1$ is 
\[\lambda_{\min}(Q_\cP^{-1}Q_\cP^c )=-\epsilon n_k\;\Rightarrow \;\lambda_{\min}(\Lambda_\cP )=1-\epsilon n_k.\]

\textit{(ii) Dynamic Partitioning.} The matrix $\Exp[Q_\cP^{-1}Q_\cP^c]$ is an expectation over the block diagonal matrices arising from different partitionings. The probability that a particular off-diagonal element is non-zero for any random partitioning is $p=n_k (K-1)/(n-1)$. This yields a matrix where the diagonal elements are zero and all off-diagonal elements are equal to $p\epsilon$. Hence, again, by symmetry, the eigenvector corresponding to the smallest eigenvalue will be $\vv=\ev_{i}-\ev_{j}$ for any $i\neq j$ and the corresponding eigenvalue is
\[\lambda_{\min}(\Exp[Q_\cP^{-1}] Q_\cP^c)= -\epsilon p\;\Rightarrow \;\lambda_{\min}(\Exp[\Lambda_\cP] )=1-\epsilon p.\] 

We conclude that for $K>1$ and $n_k>1$ we have
\[0< \lambda_{\min}(\Lambda_{\cP})\leq\lambda_{\min}(\Exp[\Lambda_\cP])\leq1\]
where the inequality is strict for any $\alpha>0$. 
Hence, repartitioning moves the smallest eigenvalue by a factor of $\frac{K-1}{n-1}\approx \frac 1 {n_k}$ closer to $1$ compared to any static partitioning. 
By inspecting $\epsilon$ we also see that the potential convergence gain of repartitioning increases as the weight $\alpha$ in the off-diagonal elements gets larger. 
This directly translates into a significantly better convergence rate as by Theorem~\ref{thm:glm}. We later verify this empirically in Section~\ref{sec:experiments}.
For an illustration of the sensitivity of the eigenvalues w.r.t $\alpha$ and $K$ we refer to Figure~\ref{fig:heatmapeiguniform} in Appendix~\ref{app:add}.

\begin{figure*}[t!]
\centering
\subfigure{\label{fig:violinsep}\includegraphics[width = 0.49\columnwidth]{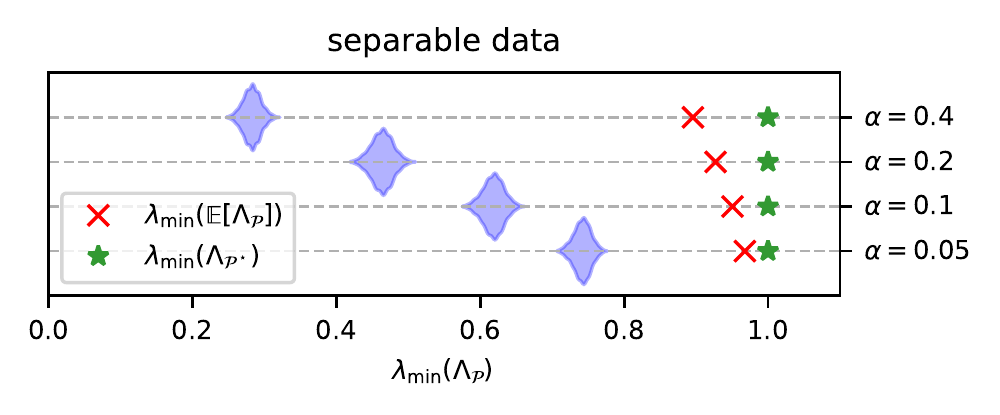}}
\subfigure{\label{fig:violinreal}\includegraphics[width = 0.49\columnwidth]{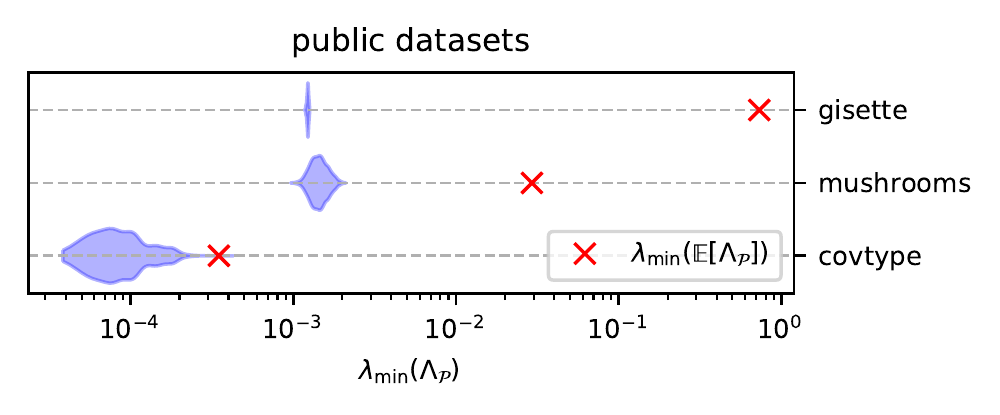}}
\vspace{-0.2cm}
\caption{ Violin plot of the distribution of $\lambda_{\min}(\Lambda_\cP)$ across 1000 random partitions on different datasets for $K=5$ partitions. We compare $\lambda_{\min}(\Lambda_\cP)$ that determines the rate of static partitioning to $\lambda_{\min}(\Exp[\Lambda_\cP])$ that governs the rate of dynamic partitioning. In the case of synthetic data where the best partitioning $\cP^\star$ is known, we also show $\lambda_{\min}(\Lambda_{\cP^\star}$).}
\label{fig:violin}
\end{figure*}

\subsection{ Separable Data}
\label{sec:separable}
Let us consider a second extreme case, where $Q$ has block diagonal structure by definition. We again assume that all non-zero off-diagonal elements are equal to $\alpha\in [0,1)$.  For $n=4, K=2$ the matrix $Q$ would look as follows:
\begin{equation*}
\setlength\arraycolsep{3pt}
Q=\left[\begin{array}{cccc}
1&\alpha&0&0\\
\alpha&1&0&0\\
0&0&1&\alpha\\
0&0&\alpha&1\\
\end{array}\right]
\end{equation*}

This could, for example, correspond to a linear regression setting where the data is perfectly separable and data columns within partitions are equally correlated.
In this case, the best static partitioning $\cP^\star$ is aligned with the block structure of the matrix. In this case $Q=Q_{\cP^\star}$, and hence $\lambda_{\min}(\Lambda_{\cP^\star})=1$. We can show that for $K>1$ it holds that 
\begin{equation}
\min_{\cP} \lambda_{\min}(\Lambda_\cP)\leq \lambda_{\min}(\Exp[\Lambda_\cP])\leq \max_\cP \lambda_{\min}(\Lambda_\cP)
\label{eq:lamsep}
\end{equation}
and equality is achieved for $\alpha=0$. Recall that the convergence rate of Algorithm \ref{alg:algo} as by Theorem \ref{thm:quad} is proportional to $\Exp[\Lambda_\cP ]$. Hence, the order in \eqref{eq:lamsep} implies that the convergence rate of repartitoning lies  between the best and the worst static partitioning. 

To investigate where on this spectrum the convergence of repartitiong actually is, we compute the distribution of $\lambda_{\min}(\Lambda_\cP)$ and the corresponding value of $\lambda_{\min}(\Exp[\Lambda_\cP ])$ numerically for different values of $\alpha$. Results are illustrated  in the left plot of Figure~\ref{fig:violin}. The violin plot suggests that even in the perfectly separable case, repartitioning achieves a significantly better convergence rate than static partitioning with probability close to 1. Hence, if we do not know the best partitioning a priori  repartitioning might be the best choice.

\subsection{Real Datasets}
\label{sec:real}
We conclude this section by considering a more practical choice of $Q$. Therefore, we consider a ridge regression problem with $Q=A^\top A+\lambda I$. We choose $\lambda=1$ and we evaluate $\lambda_{\min}(\Lambda_\cP)$ numerically for some popular datasets. We have chosen the gisette, the mushroom and the covtype dataset that can be downloaded from \citep{ucirepo} and whose statistics are reported in Table~\ref{tb:datasets}. 

In the right plot of Figure~\ref{fig:violin} we compare the  distribution of $\lambda_{\min}(\Lambda_\cP)$ for the three datsets across random partitionings $\cP$ with $\lambda_{\min}(\Exp[\Lambda_\cP])$. We see that across all datasets $\rho$ for random partitioning is higher than for any fixed partitioning with very high probability which implies superior convergence of repartitioning as by our theory. This observation is also consistent across different choices of regularizer as shown in Figure~\ref{fig:violinapp} in the appendix for completeness.

\begin{figure*}[t]
\centering
\subfigure{\label{fig:hist_ex2}\footnotesize
\begin{tabular}{ c|c|c} 
$\alpha = 0.1$ &$\frac 1 K \lambda_{\min}(\Exp[\Lambda_\cP])$&$\frac 1 K \lambda_{\min}(\Lambda_\cP)$  \\ \hline
 $ K=2$& 0.498 & 0.040  \\ 
 $ K=4$ & 0.247 & 0.038\\ 
$ K=8$& 0.122& 0.034\\
\end{tabular}}
\hspace{0.7cm}
\subfigure{\label{fig:hist_ex2}\footnotesize
\begin{tabular}{ c|c|c} 
 $K=5$&$\frac 1 K \lambda_{\min}(\Exp[\Lambda_\cP])$&$\frac 1 K \lambda_{\min}(\Lambda_\cP)$  \\ \hline
 $\alpha = 0.01$& 0.199 & 0.142 \\ 
 $\alpha = 0.1$ & 0.197& 0.037\\ 
$\alpha = 0.5$&0.196 &0.005\\
\end{tabular}}
 \addtocounter{subfigure}{-2}
\subfigure[sensitivity w.r.t. $K$ ($\alpha=0.1$)]{\label{fig:hist_ex2}\includegraphics[width=0.45\columnwidth]{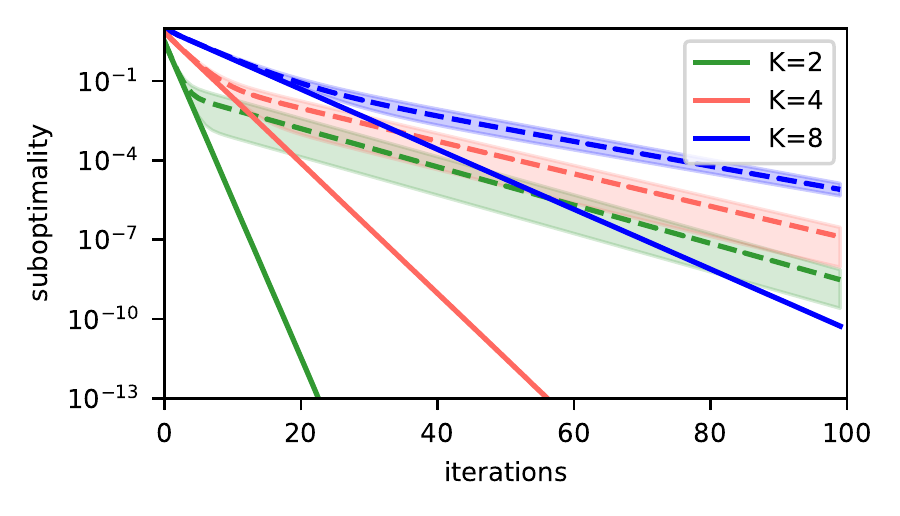}}\hspace{0.5cm}
\subfigure[sensitivity w.r.t. $\alpha$ ($K=5$)]{\label{fig:hist_ex2}\includegraphics[width=0.45\columnwidth]{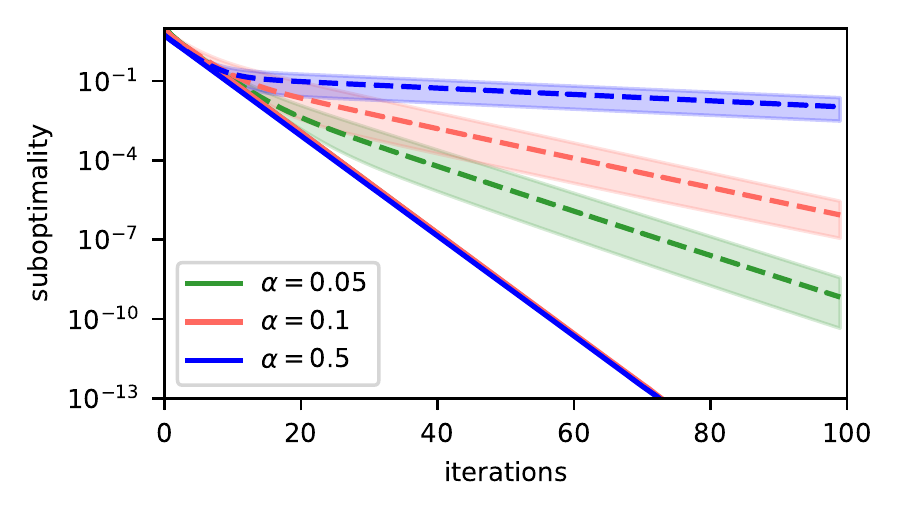}}
\caption{Linear regression on synthetic data ($n=200$) with uniform correlations of strength $\alpha$. We compare the empirical  convergence of Algorithm~\ref{alg:algo} for static (dashed) and dynamic (solid) partitioning to the corresponding theoretical convergence rate $\rho = \frac 1 K  \lambda_{\min}(\Exp[\Lambda_\cP]) $, see Theorem~\ref{thm:quad}, for different values of $K$ and $\alpha$. Confidence intervals show min-max intervals over 100 runs.}
\label{fig:perfuniform}
\end{figure*}

\section{Performance Results}
\label{sec:experiments}

Finally, we compare the convergence gains of repartitioning predicted by our theory, with the actual convergence  of Algorithm \ref{alg:algo} with and without  repartitioning. We consider two popular machine learning problems. First, we consider linear regression where 
\[f(\xv)=\frac 1 2 \|A\xv-\yv\|^2 + \frac \lambda 2 \|\xv\|^2\] 
and the second-order model $\tilde f_\xv$ is exact with $Q_t=A^\top A+ \lambda I $. This allows us to analyse the scenarios discussed in Section~\ref{sec:eff} and the gains predicted by Theorem~\ref{thm:quad}. 
As a second application we consider $L_2$-regularized logistic regression with \[f(\xv)=\sum_{i\in[n]} \log\left(1 + \exp(-y_i A_{i,:}\xv)\right) + \frac \lambda 2 \|\xv\|^2\]
for $y_i \in\{\pm 1\}$ where we use the second-order Taylor expansion for defining $\tilde f_\xv(\argdot,Q_t)$. This corresponds to the general case analyzed in Theorem~\ref{thm:general} where $Q_t$ depends on the model $\xvt$ and changes across iterations.

\subsection{Validation of Convergence Rates}

Let us revisit the synthetic examples from Section~\ref{sec:eff} and verify the convergence of Algorithm~\ref{alg:algo} empirically. We start with the uniform correlation example from Section~\ref{sec:uniform} and generate a synthetic data matrix $A=Q^{1/2}$ together with random labels $\yv$. We then train a linear regression model and investigate the convergence of Algorithm~\ref{alg:algo} for (i) static and (ii) dynamic partitioning. In Figure~\ref{fig:perfuniform} we contrast the convergence results to the theoretical rate predicted by Theorem~\ref{thm:quad} which we can evaluate using the expressions derived in Section~\ref{sec:uniform}. We perform this experiment for different values of $K$ and $\alpha$. The tables on the top contain the values of the convergence rate $\rho= \frac 1 K  \lambda_{\min}(\Exp[\Lambda_\cP])$  for the corresponding figures at the bottom. We observe a very close match between the relative gain of repartitioning over static partitioning predicted by the theory and the empirical behavior. This supports that $\lambda_{\min}(\Exp[\Lambda_\cP])$ indeed captures the effect of repartitioning accurately.

\begin{figure*}[t!]
\centering
\subfigure{\includegraphics[width=0.32\columnwidth]{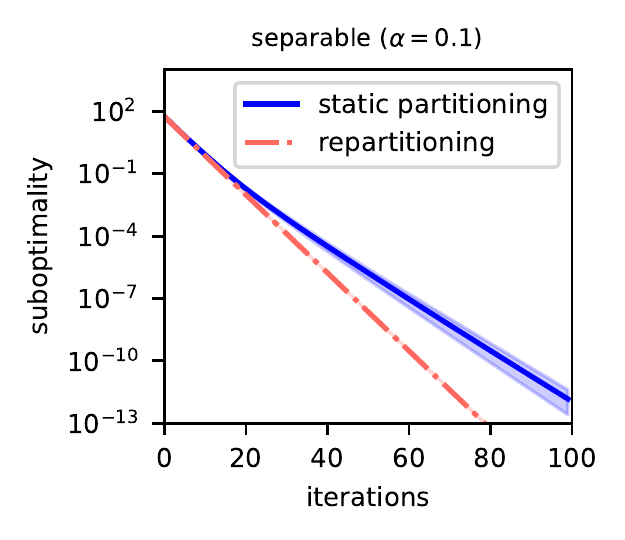}}
\subfigure{\includegraphics[width=0.32\columnwidth]{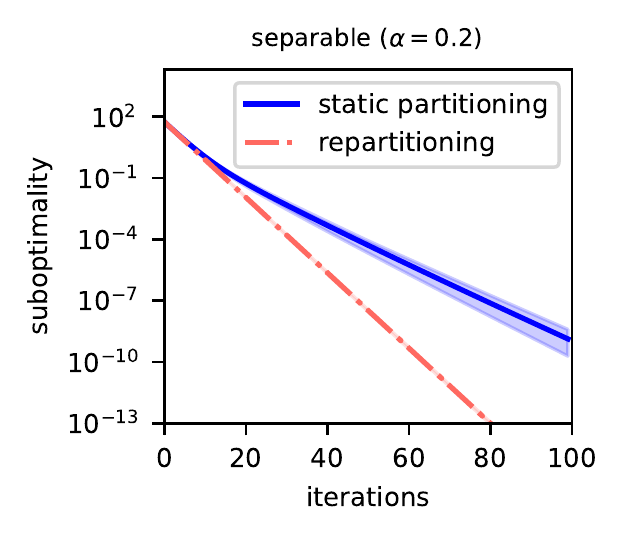}}
\subfigure{\includegraphics[width=0.32\columnwidth]{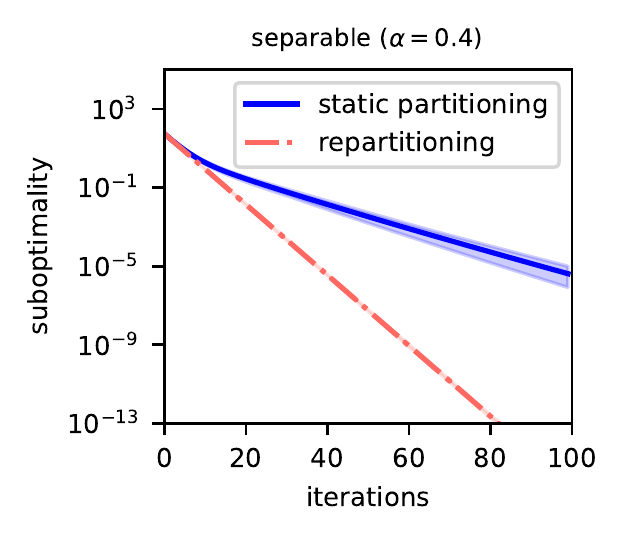}}
\subfigure{\includegraphics[width=0.33\columnwidth]{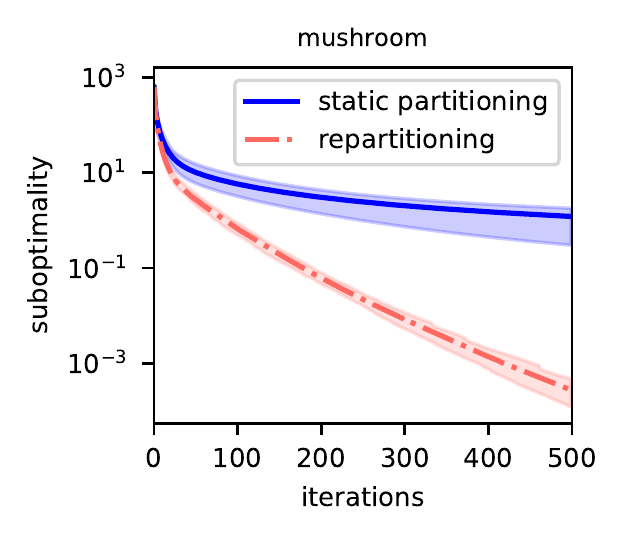}}
\subfigure{\includegraphics[width=0.33\columnwidth]{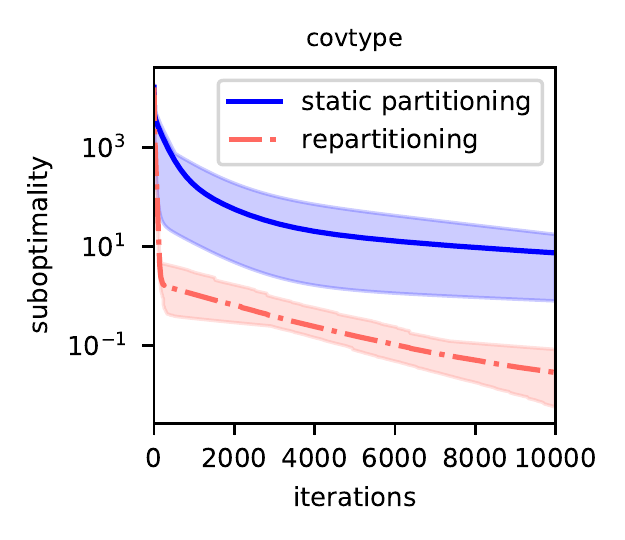}}
\vspace{-0.2cm}
\caption{Empirical performance of Algorithm~\ref{alg:algo} for linear regression on a selection of the datasets analyzed in Figure~\ref{fig:violin}.  The relative convergence of Alg~\ref{alg:algo} with and without repartitioning closely match the values predicted by our theory as given through $\lambda_{\min}(\Lambda_\cP)$  (see Theorem~\ref{thm:quad}) whose values are illustrated for the respective datasets  in Figure~\ref{fig:violin}. Confidence intervals show min-max intervals over 10 runs.}
\label{fig:evhist-conv}
\end{figure*}

\begin{table}
\centering
\small
\begin{tabular}{c | c | c}
Dataset & \# datapoints & \# features \\
\hline
mushroom & 8124 & 112 \\
covtype & 581012 & 54 \\
gisette & 6000 & 5000 \\
rcv1 & 20’242 & 677399 \\
url & 2396130 & 3231961 \\
synthetic & 200 & 200 \\
\end{tabular}
\caption{Size of the datasets used in our experimental results.}
\label{tb:datasets}
\end{table}

We further verify the empirical convergence for the separable data from Section~\ref{sec:separable} as well as the ridge regression setting from Section~\ref{sec:real}.
The convergence results of Algorithm~\ref{alg:algo} are depicted in Figure~\ref{fig:evhist-conv} for a subset of the parameter settings. 
Again, we observe a strong correlation between the empirical convergence gains and the values of $\lambda_{\min}(\Lambda_\cP)$ evaluated numerically in Figure~\ref{fig:violin} across all datasets.

To be consistent with the assumptions used in our theorems, we have implemented Algorithm~\ref{alg:algo} with a fixed step size $\eta$. 
Alternatively, the algorithm could also be augmented with backtracking line search~\citep{armijo1966}. This would potentially improve the performance of good partitionings even further, but it is not expected to significantly change the relative behavior of static versus dynamic partitioning which is the main study of this paper. To support this claim we compare the convergence of  Algorithm \ref{alg:algo} for the two partitioning schemes with and without backtracking line-search on three synthetic examples in Figure~\ref{fig:ls} in the appendix.

\subsection{Existing Algorithms}
To complete our study we have implemented three popular existing distributed methods and combined them with repartitioning. These are \cocoa \citep{Smith:2016wp} with a dual solver, ADN \citep{duenner2018adn} and the line-search-based approach \citep{lee2017distributed}, referred to as LS. We have trained a logistic regression classifier using all three algorithms on three different datasets and illustrate the respective convergence with and without repartitioning in Figure~\ref{fig:perf}. Additional results for ADN on two more datasets can be found in Figure \ref{fig:ADNapp} in the appendix.  Overall, we see a consistent and significant gain of repartitioning for all three methods. We find that the potential convergence gain  mostly depends on the statistics of the datasets (which defines $Q_t$) and is similar for all methods. For the url data repartitioning with a dual solver reduces sample complexity by several orders of magnitude, for gisette the gain is around $30\times$ and for the rcv1 dataset it is $2\times$. When inspecting the properties of the datasets we find that the gain of randomization grows with the density and the dimension of the columns of the data matrix $A$. This is expected, because it implies stronger correlations and hence more weight in the off-diagonal elements of $Q_t$.

\begin{figure*}[t]
\centering
\subfigure[CoCoA-dual \citep{Smith:2016wp}]{\label{fig:cocoa}\includegraphics[width=0.32\columnwidth]{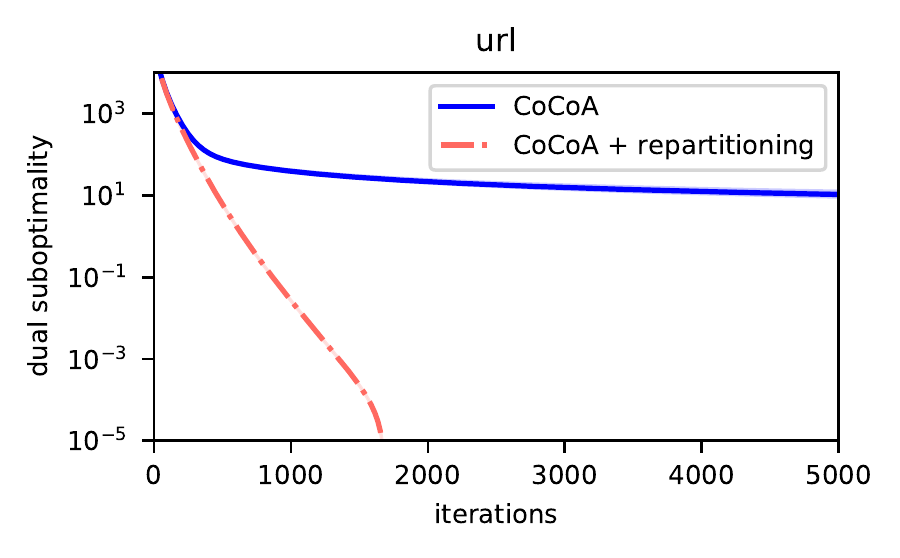}}
\subfigure[LS \citep{lee2017distributed}]{\label{fig:ls}\includegraphics[width=0.32\columnwidth]{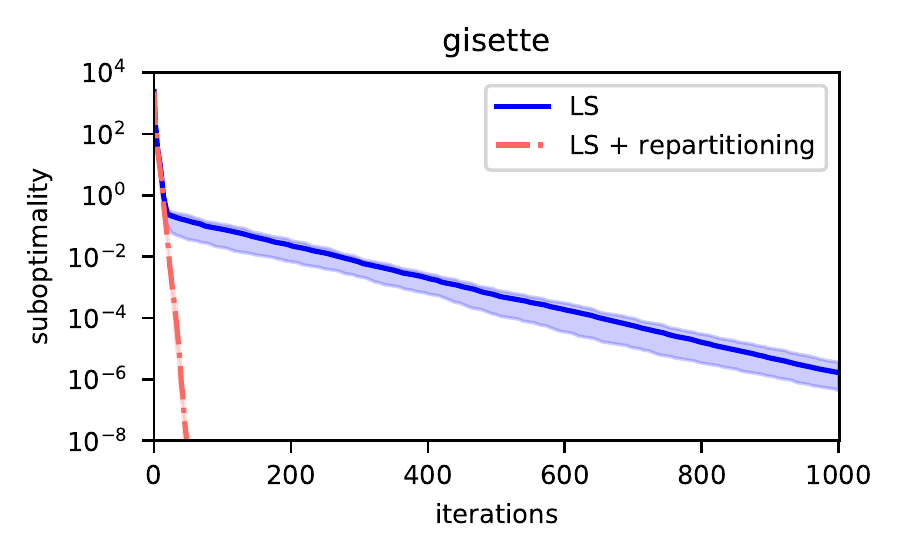}}
\subfigure[ADN \citep{duenner2018adn}]{\label{fig:TR}\includegraphics[width=0.32\columnwidth]{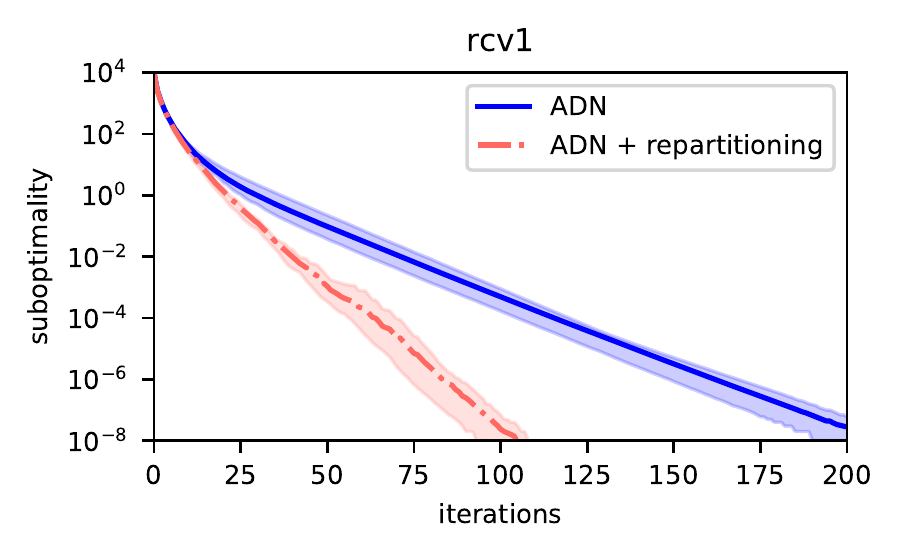}}
\caption{Combining existing distributed methods with repartitioning: Training of a logistic regression classifier with $\lambda = 1$ using three different algorithms and datasets for $K=8$. Confidence intervals show min-max intervals over 10 runs. Experiments on additional datasets and different values of $K$ can be found in Figure~\ref{fig:ADNapp} in the appendix. }
\label{fig:perf}
\end{figure*}

\section{Conclusion}
We have demonstrated theoretically, as well as empirically, that repartitioning can improve the sample complexity of existing block-separable optimization methods by potentially several orders of magnitude. The gain crucially depends on the problem at hand and is accurately captured by a simple analytical quantity identified in our theoretical analysis.

Together with prior work \citep{syscd} that demonstrated the implementation efficiency of block-separable models on modern hardware, our results highlight that repartitioning can turn existing distributed methods into promising candidates for parallel learning. An additional important benefit of these methods is that they come with convergence guarantees for arbitrary degrees of parallelism without prior assumptions on the data. This allows them to be scaled to any number of available cores.

Finally, we would like to note that the repartitioning technique discussed in this manuscript is versatile and our analysis is intentionally kept general to cover different types of algorithms and preconditioning matrices.

\section*{Acknowledgements}
The authors would like to thank Foivos Alimisis for proofreading our analysis and suggesting an improvement to Theorem~\ref{thm:glm}.
Further, we wish to acknowledge support from the Swiss National Science Foundation Early Postdoc Mobility Fellowship Program.

\bibliography{references}
\bibliographystyle{plainnat}

\newpage
\onecolumn

\appendix

\section{Convergence Proof of Section \ref{sec:conv}}

\subsection{Proof Theorem \ref{thm:quad}}

For a quadratic function $f$ as in \eqref{eq:fquad} the second-order Taylor expansion is exact and
\begin{equation}f(\xv+\Delta\xv)=f(\xv) +\nabla f(\xv)^\top \Delta \xv+ \frac 1 2 \Delta\xv^\top H\Delta\xv.
\label{eq:squarefp}
\end{equation}
Furthermore, the update $\Delta \xv^\star=-H^{-1}\nabla f(\xv)$ is optimal, in the sense that $f(\xv+\Delta \xv^\star)=f^\star$. We will now analyze the update step $\xvtp =\xvt +\Delta \xv$ of Algorithm~\ref{alg:algo} with $\Delta\xv = - \eta H_\cPt^{-1} \nabla f(\xvt)$. Recall that $H_\cPt$ denotes the block diagonal version of $H$ induced by the partitioning $\cPt$ at iteration $t$. Plugging in the update we find
\begin{align}
f(\xvtp)-f^\star&=f(\xvt - \eta H_\cPt^{-1} \nabla f(\xvt))-f(\xvt-H^{-1}\nabla f(\xvt)) \nonumber\\
&\overset{\eqref{eq:squarefp}}{=} \nabla f(\xvt)^\top \left(  H^{-1}-\eta H_\cPt^{-1}\right) \nabla f(\xvt) - \frac 1 2 \nabla f(\xvt)^\top [ H^{-\top} H H^{-1}-\eta^2 H_\cPt^{-\top} H H_\cPt^{-1}] \nabla f(\xvt)
\end{align}
Further, using the bound $\xv^\top H \xv\leq K \xv^\top H_\cPt \xv$ which holds for all $\xv$ we find
\begin{align}
f(\xvtp)-f^\star&= \nabla f(\xvt)^\top \left(  -\eta H_\cPt^{-1} + \frac 1 2 H^{-1}+\frac{\eta^2} 2 H_\cPt^{-\top} H H_\cPt^{-1} \right) \nabla f(\xvt)\nonumber\\
&\leq \nabla f(\xvt)^\top \left(  -\eta H_\cPt^{-1} + \frac 1 2 H^{-1}+ K \frac{\eta^2} 2 H_\cPt^{-\top} H_\cPt H_\cPt^{-1}\right) \nabla f(\xvt)\nonumber\\
&= \nabla f(\xvt)^\top \left(  \frac 1 2 H^{-1}+ [K \frac{\eta^2} 2-\eta] H_\cPt^{-1}\right) \nabla f(\xvt)\nonumber\\
&= \frac 1 2 \nabla f(\xvt)^\top  \left(  I+ [K {\eta^2}-2 \eta] H_\cPt^{-1} H\right) H^{-1} \nabla f(\xvt)
\end{align}
Now, using the fact that $\nabla f(\xvs)=0$ and plugging in the exact expression of the gradient ($\nabla f(\xv)=H\xv -\cv$) and the Hessian ($\nabla^2 f(\xv)=H$) we find
\begin{align}
f(\xvtp)-f^\star &\leq\frac 1 2 (\nabla f(\xvt)-\nabla f(\xvs))^\top  \left(  I+ [K {\eta^2}-2 \eta] H_\cPt^{-1} H\right) H^{-1} (\nabla f(\xvt)-\nabla f(\xvs))\nonumber\\
&= \frac 1 2 (\xvt-\xvs)^\top H  \left(  I+ [K {\eta^2}-2 \eta] H_\cPt^{-1} H\right) (\xvt-\xvs)\nonumber
\end{align}
Now, using $\eta = \frac 1 K$ and take expectation w.r.t. the randomness of the partitioning we find
\begin{align}
\Exp[f(\xvtp)-f^\star] &\leq  \frac 1 2 (\xvt-\xvs)^\top H  \left(  I-\frac 1 K \Exp[H_\cPt^{-1}] H\right) (\xvt-\xvs)\nonumber\\
&\overset{(i)} \leq\frac 1 2 \lambda_{\max}\left(  I - \frac 1 K \Exp[H_\cPt^{-1}] H\right)(\xvt-\xvs)^\top H   (\xvt-\xvs)\nonumber\\
&=\frac 1 2 \left[ 1- \frac 1 K \lambda_{\min}(  \Exp[H_\cPt^{-1}] H)\right](\xvt-\xvs)^\top H   (\xvt-\xvs)\nonumber\\
&=   \left[ 1-  \frac 1 K \lambda_{\min}(  \Exp[H_\cPt^{-1}] H)\right]\left(f(\xvt)-f(\xvs)\right).\label{eq:ratequadf}
\end{align}
which concludes the proof. Note that in (i) we used the fact that $H$ is symmetric and PSD. The last equality followed from the definition of $f$ in  \eqref{eq:fquad}:
\begin{align}
(\xvt-\xvs)^\top H   (\xvt-\xvs) &=\xvt ^\top H \xvt +\xvs^\top H \xvs - 2 \xvt^\top H \xvs\nonumber\\
&=\xvt ^\top H \xvt \pm 2 \cv^\top \xvt +\xvs^\top H \xvs \pm 2 \cv^\top \xvs - 2 \xvt^\top H \xvs\nonumber\\
&=2 f(\xvt) +\xvs^\top H \xvs \pm 2 \cv^\top \xvs \nonumber\\
&=2 f(\xvt) - 2f(\xvs) + 2\xvs^\top H \xvs - 2 \cv^\top \xvs \nonumber\\
&=2( f(\xvt) -f(\xvs)).
\end{align}

\subsection{Proof of Lemma \ref{lem:fdecglm}}
\label{app:lemglm}

We are given the  quadratic auxiliary model \eqref{eq:objglm} and $\nabla^2 f(\xvt)=A^\top \nabla^2\ell(A\xvt) A\preccurlyeq\gamma_ \ell A^\top A$, we have
\begin{align}
f(\xvt + \Delta \xv)\leq f(\xvt) + \nabla f(\xvt)^\top \Delta \xv + \frac {\gamma_\ell} 2 \Delta\xv^\top M\Delta\xv 
\label{eq:modelM}
\end{align}
where $M:=A^\top A$ is a fixed symmetric matrix. Recall that we use the notation $M_\cPt$ to denote the block diagonal version of $M$ induced by the partitioning $\cPt$ at iteration $t$. Using the block-diagonal matrix as preconditioning matrix in the update step of Algorithm~\ref{alg:algo} we have 
\[\xvtp = \xvt - \eta M_\cPt^{-1} \nabla f(\xvt)\] 
Plugging $\Dxv= - \eta M_\cPt^{-1} \nabla f(\xvt)$ into the auxiliary model \eqref{eq:modelM} yields
\begin{align}
f(\xvtp)&\leq f(\xvt)  - \eta \nabla f(\xvt)^\top  M_\cPt^{-1} \nabla f(\xvt)  + \eta^2 \frac { \gamma_\ell} 2  \nabla f(\xvt)^\top  M_\cPt^{-1} M M_\cPt^{-1} \nabla f(\xvt)\nonumber\\
&\overset{(i)}\leq f(\xvt)  - \eta \nabla f(\xvt)^\top  M_\cPt^{-1} \nabla f(\xvt)  + K \eta^2 \frac {\gamma_\ell} {2}  \nabla f(\xvt)^\top  M_\cPt^{-1} M_\cPt M_\cPt^{-1} \nabla f(\xvt)\nonumber\\
&= f(\xvt)  - \eta \nabla f(\xvt)^\top  M_\cPt^{-1} \nabla f(\xvt)  + K \eta^2 \frac {\gamma_\ell} {2} \nabla f(\xvt)^\top M_\cPt^{-1}  \nabla f(\xvt)\nonumber\\
&= f(\xvt)  -  \left[ \eta  - K \eta^2 \frac {\gamma_\ell} {2} \right]  \nabla f(\xvt)^\top  M_\cPt^{-1}  \nabla f(\xvt)
\end{align}
where we used the inequality  $\xv^\top M\xv\leq K \xv^\top M_\cPt \xv\,\,\forall \xv$ in $(i)$.
Now, subtracting $f(\xvt)$ on both sides, using the step size $\eta = \frac 1 {K\gamma_\ell}$ and changing signs we get
\begin{align}
f(\xvt)-f(\xvtp)&\geq   \frac 1 {2K \gamma_\ell} \nabla f(\xvt)^\top  M_\cPt^{-1}  \nabla f(\xvt)\nonumber\\
&=   \frac 1 {2K \gamma_\ell}  \nabla \ell(A\xvt)^\top A  M_\cPt^{-1}A^\top  \nabla \ell( A\xvt)
\end{align}
We can now take expectations w.r.t. the randomness of the partitioning on both sides which concludes the proof of Lemma \ref{lem:fdecglm},
\begin{align}
f(\xvt)-\Exp[f(\xvtp)]&\geq   \frac 1 {2K\gamma_\ell}  \nabla \ell(A\xvt)^\top A  \Exp[M_\cPt^{-1}]A^{\top}  \nabla \ell( A\xvt)\nonumber\\
&\geq     \frac 1 {2K\gamma_\ell} \lambda_{\min}(A \Exp[M_\cPt^{-1}]A^{\top}) \| \nabla \ell(A\xvt)\|^2. \label{eq:lem2}
\end{align}

\subsection{Proof of Theorem \ref{thm:glm}}
\label{app:thmglm}

Given that  $\ell$ satisfies the Polyak-Lojasiewicz inequality \eqref{eq:PL} we can lower bound the gradient norm in \eqref{eq:lem2} by the suboptimality which concludes the proof of Theorem \ref{thm:glm}.

\subsection{Proof of Lemma \ref{lem:general}}
\label{app:lemgeneral}

We assume the second-order model $\tilde f_\xv$ satisfies
\begin{align}
f(\xvt + \Delta \xv)&\leq \xi \hat f_{\xvt}(\Delta\xv, Q_t) + (1-\xi) f(\xvt)\notag\\
&\leq\xi [ f(\xvt) + \nabla f(\xvt)^\top \Delta \xv + \frac 1 2 \Delta\xv^\top Q_t\Delta\xv] + (1-\xi) f(\xvt)\notag\\
&\leq  f(\xvt) +  \xi[\nabla f(\xvt)^\top \Delta \xv + \frac 1 2 \Delta\xv^\top Q_t\Delta\xv] 
\label{eq:e1}
\end{align}
Following the notation of the paper we will denote the  block diagonal  version of $Q_t$ by $Q_\cPt$ and use this as preconditioning matrix. Plugging the update step $\Dxv= - \eta Q_\cPt^{-1} \nabla f(\xvt)$ into \eqref{eq:e1} yields
\begin{align}
f(\xvtp)\leq f(\xvt)  - \xi\eta \nabla f(\xvt)^\top  Q_\cPt^{-1} \nabla f(\xvt)  +\xi\frac 1 2 \eta^2 \nabla f(\xvt)^\top  Q_\cPt^{-1} Q_t Q_\cPt^{-1} \nabla f(\xvt)
\end{align}
we can further use that $Q_t\preccurlyeq K Q_\cPt$ which holds for every $\cPt$ and thus 
\begin{align}
f(\xvtp)&\leq f(\xvt)  - \xi\eta \nabla f(\xvt)^\top  Q_\cPt^{-1} \nabla f(\xvt)  + \xi\frac K {2} \eta^2 \nabla f(\xvt)^\top  Q_\cPt^{-1} Q_\cPt Q_\cPt^{-1} \nabla f(\xvt)\notag\\
&= f(\xvt)  -\xi \eta \nabla f(\xvt)^\top  Q_\cPt^{-1} \nabla f(\xvt)  +\xi \frac K {2} \eta^2 \nabla f(\xvt)^\top  Q_\cPt^{-1}  \nabla f(\xvt)\notag\\
&= f(\xvt)  -  \xi\left[ \eta  - \frac K {2} \eta^2\right]  \nabla f(\xvt)^\top  Q_\cPt^{-1}  \nabla f(\xvt).
\end{align}
Plugging in the step size $\eta=\frac 1 K$, subtracting $f(\xvt)$ on both sides and changing signs we get
\begin{align}
f(\xvt)-f(\xvtp)&\geq  \xi \frac 1 {2K} \nabla f(\xvt)^\top  Q_\cPt^{-1}  \nabla f(\xvt)
\end{align}

Now let us use the fact that 
\begin{align}
\nabla_\xvt  \tilde f_{\xvt}(\xvtp-\xvt, Q_t) &= \nabla f(\xvt) + Q_t (\xvtp-\xvt) \\
\nabla_\xvt  \tilde f_{\xvt}(0,Q_t) &= \nabla f(\xvt)
\end{align} 
and define $\Delta \tilde \xvt^\star=\argmin_\xv \tilde f_{\xvt}(\xv)  $ to be the optimizer of the quadratic approximation \eqref{eq:obj2} around $\xvt$. This yields
\begin{align}
f(\xvt)-f(\xvtp)&\geq  \xi   \frac 1 {2K} \left[\nabla_\xvt  \tilde f_{\xvt}(0,Q_t)-\nabla_\xvt  \tilde f_{\xvt}(\Delta \tilde\xvt^\star, Q_t)   \right]^\top  Q_\cPt^{-1}  \left[\nabla_\xvt  \tilde f_{\xvt}(0,Q_t)-\nabla_\xvt  \tilde f_{\xvt}(\Delta \tilde \xvt^\star, Q_t)   \right]\notag\\
&=    \xi \frac 1 {2K} \left[Q_t(\xvt - \tilde \xvt^\star ) \right]^\top  Q_\cPt^{-1}  \left[ Q_t(\xvt - \tilde \xvt^\star )   \right]\notag\\
&=   \xi \frac 1 {2K} (\xvt - \tilde \xvt^\star )^\top  Q_t^\top  Q_\cPt^{-1}  Q_t(\xvt - \tilde \xvt^\star )  \label{eq:b}.
\end{align}
Now, taking expectations w.r.t, the randomness in the partitioning we get the expression from Lemma~\ref{lem:general}:
\begin{align}
\Exp[f(\xvt)-f(\xvtp)]&\geq \xi \frac 1 {2K} (\xvt - \tilde \xvt^\star )^\top  Q_t^\top \Exp[  Q_\cPt^{-1} ] Q_t(\xvt - \tilde \xvt^\star ) \notag \\
&\geq\xi {\frac 1 {2K} \lambda_{\min}( Q_t^\top \Exp[Q_\cPt^{-1}]Q_t  )}  \|\xvt - \tilde \xvt^\star\|^2\label{eq:eq2}.
\end{align}


\section{Proof of Theorem \ref{thm:general}}

Using the assumption from \eqref{ass:alpha} which states
\[f(\tilde \xvt)\leq\alpha f(\xvt)+ (1-\alpha) f(\xvs)\]
we can relate $  \|\xvt - \tilde \xvt^\star\|^2$ to the suboptimality as follows:
\begin{align*}
f(\xvt) - f^\star &= f(\xvt) - f(\tilde \xvt^\star) + f(\tilde \xvt^\star) - f^\star \nonumber \\
& \leq L \| \xvt - \tilde \xvt^\star \| + \alpha (f(\xvt) - f^\star) \\
&\\
\implies& (1-\alpha) (f(\xvt) - f^\star) \leq L \| \xvt - \tilde \xvt^\star \|
\end{align*}
where we used the fact that $f$ is $L$-Lipschitz continuous.\\
Now denoting 
\[\xi_t:=\xi\frac 1 {2K}  \lambda_{\min}(Q_t^\top \Exp[Q_\cPt^{-1}]Q_t  )\] 
and going from \eqref{eq:eq2} we have
\begin{align*}
\Exp [ f(\xvtp) - f(\xvt)] &\leq  -\xi_t \frac{(1-\alpha)}{L} (f(\xvt) - f^\star) \\
\implies \Exp f(\xvtp) - f^\star &\leq (f(\xvt) - f^\star) - \xi_t \frac{(1-\alpha)}{L} (f(\xvt) - f^\star) \\
&= \left[1 - \xi_t \frac{(1-\alpha)}{L} \right] (f(\xvt) - f^\star).
\end{align*}
Theorem \ref{thm:general} follows by unrolling the recursion:
\begin{align}
\Exp f(\xvtp) - f^\star &\leq \left[1 - \min_t \xi_t \frac{(1-\alpha)}{L} \right]^t (f(\xv_0) - f^\star).
\end{align}

\newpage
\section{Spectral Analysis}

\subsection{Uniform Correlation} 
\label{app:uniform}

Let us visualize the matrices involved in the uniform-data example discussed in Section~\ref{sec:uniform} for $n=9$, $K=3$, $n_k=3$:

\begin{figure*}[h!]
\small
\begin{equation*}
\setlength\arraycolsep{3.7pt}
\underbrace{\left[\begin{array}{ccccccccc}
\redcell1&\redcell\alpha&\redcell\alpha&\redcell\alpha&\redcell\alpha&\redcell\alpha&\redcell\alpha&\redcell\alpha&\redcell\alpha\\
\redcell\alpha&\redcell1&\redcell\alpha&\redcell\alpha&\redcell\alpha&\redcell\alpha&\redcell\alpha&\redcell\alpha&\redcell\alpha\\
\redcell\alpha&\redcell\alpha&\redcell1&\redcell\alpha&\redcell\alpha&\redcell\alpha&\redcell\alpha&\redcell\alpha&\redcell\alpha\\
\redcell\alpha&\redcell\alpha&\redcell\alpha&\redcell1&\redcell\alpha&\redcell\alpha&\redcell\alpha&\redcell\alpha&\redcell\alpha\\
\redcell\alpha&\redcell\alpha&\redcell\alpha&\redcell\alpha&\redcell1&\redcell\alpha&\redcell\alpha&\redcell\alpha&\redcell\alpha\\
\redcell\alpha&\redcell\alpha&\redcell\alpha&\redcell\alpha&\redcell\alpha&\redcell1&\redcell\alpha&\redcell\alpha&\redcell\alpha\\
\redcell\alpha&\redcell\alpha&\redcell\alpha&\redcell\alpha&\redcell\alpha&\redcell\alpha&\redcell1&\redcell\alpha&\redcell\alpha\\
\redcell\alpha&\redcell\alpha&\redcell\alpha&\redcell\alpha&\redcell\alpha&\redcell\alpha&\redcell\alpha&\redcell1&\redcell\alpha\\
\redcell\alpha&\redcell\alpha&\redcell\alpha&\redcell\alpha&\redcell\alpha&\redcell\alpha&\redcell\alpha&\redcell\alpha&\redcell1\\
\end{array}\right]}_{\textstyle Q}
\quad
\underbrace{\left[\begin{array}{ccccccccc}
\redcell1&\redcell\alpha&\redcell\alpha&0&0&0&0&0&0\\
\redcell\alpha&\redcell1&\redcell\alpha&0&0&0&0&0&0\\
\redcell\alpha&\redcell\alpha&\redcell 1&0&0&0&0&0&0\\
0&0&0&\redcell1&\redcell\alpha&\redcell\alpha&0&0&0\\
0&0&0&\redcell\alpha&\redcell1&\redcell\alpha&0&0&0\\
0&0&0&\redcell\alpha&\redcell\alpha&\redcell1&0&0&0\\
0&0&0&0&0&0&\redcell1&\redcell\alpha&\redcell\alpha\\
0&0&0&0&0&0&\redcell\alpha&\redcell1&\redcell\alpha\\
0&0&0&0&0&0&\redcell\alpha&\redcell\alpha&\redcell1\\
\end{array}\right]}_{\textstyle Q_\cP}
\quad
\underbrace{\left[\begin{array}{ccccccccc}
0&0&0&\redcell\alpha&\redcell\alpha&\redcell\alpha&\redcell\alpha&\redcell\alpha&\redcell\alpha\\
0&0&0&\redcell\alpha&\redcell\alpha&\redcell\alpha&\redcell\alpha&\redcell\alpha&\redcell\alpha\\
0&0&0&\redcell\alpha&\redcell\alpha&\redcell\alpha&\redcell\alpha&\redcell\alpha&\redcell\alpha\\
\redcell\alpha&\redcell\alpha&\redcell\alpha&0&0&0&\redcell\alpha&\redcell\alpha&\redcell\alpha\\
\redcell\alpha&\redcell\alpha&\redcell\alpha&0&0&0&\redcell\alpha&\redcell\alpha&\redcell\alpha\\
\redcell\alpha&\redcell\alpha&\redcell\alpha&0&0&0&\redcell\alpha&\redcell\alpha&\redcell\alpha\\\redcell\alpha&\redcell\alpha&\redcell\alpha&\redcell\alpha&\redcell\alpha&\redcell\alpha&0&0&0\\\redcell\alpha&\redcell\alpha&\redcell\alpha&\redcell\alpha&\redcell\alpha&\redcell\alpha&0&0&0\\
\redcell\alpha&\redcell\alpha&\redcell\alpha&\redcell\alpha&\redcell\alpha&\redcell\alpha&0&0&0\\
\end{array}\right]}_{\textstyle Q_\cP^c}
\end{equation*}\\
\begin{equation*}
\setlength\arraycolsep{3.7pt}
\underbrace{\left[\begin{array}{ccccccccc}
\greencell c&\greencell\beta&\greencell\beta&0&0&0&0&0&0\\
\greencell\beta&\greencell c&\greencell\beta&0&0&0&0&0&0\\
\greencell\beta&\greencell\beta&\greencell c&0&0&0&0&0&0\\
0&0&0&\greencell c&\greencell\beta&\greencell\beta&0&0&0\\
0&0&0&\greencell\beta&\greencell c&\greencell\beta&0&0&0\\
0&0&0&\greencell\beta&\greencell\beta&\greencell c &0&0&0\\
0&0&0&0&0&0&\greencell c&\greencell\beta&\greencell\beta\\
0&0&0&0&0&0&\greencell\beta&\greencell c&\greencell\beta\\
0&0&0&0&0&0&\greencell\beta&\greencell\beta&\greencell c\\
\end{array}\right]}_{\textstyle Q_\cP^{-1}}
\quad
\underbrace{\left[\begin{array}{ccccccccc}
0&0&0&\bluecell\epsilon&\bluecell\epsilon&\bluecell\epsilon&\bluecell\epsilon&\bluecell\epsilon&\bluecell\epsilon\\
0&0&0&\bluecell\epsilon&\bluecell\epsilon&\bluecell\epsilon&\bluecell\epsilon&\bluecell\epsilon&\bluecell\epsilon\\
0&0&0&\bluecell\epsilon&\bluecell\epsilon&\bluecell\epsilon&\bluecell\epsilon&\bluecell\epsilon&\bluecell\epsilon\\
\bluecell\epsilon&\bluecell\epsilon&\bluecell\epsilon&0&0&0&\bluecell\epsilon&\bluecell\epsilon&\bluecell\epsilon\\
\bluecell\epsilon&\bluecell\epsilon&\bluecell\epsilon&0&0&0&\bluecell\epsilon&\bluecell\epsilon&\bluecell\epsilon\\
\bluecell\epsilon&\bluecell\epsilon&\bluecell\epsilon&0&0&0&\bluecell\epsilon&\bluecell\epsilon&\bluecell\epsilon\\
\bluecell\epsilon&\bluecell\epsilon&\bluecell\epsilon&\bluecell\epsilon&\bluecell\epsilon&\bluecell\epsilon&0&0&0\\
\bluecell\epsilon&\bluecell\epsilon&\bluecell\epsilon&\bluecell\epsilon&\bluecell\epsilon&\bluecell\epsilon&0&0&0\\
\bluecell\epsilon&\bluecell\epsilon&\bluecell\epsilon&\bluecell\epsilon&\bluecell\epsilon&\bluecell\epsilon&0&0&0\\
\end{array}\right]}_{\textstyle Q_\cP^{-1}Q_\cP^c }
\quad
 \underbrace{ \left[\begin{array}{ccccccccc}
0&p \graycell\epsilon&p \graycell\epsilon&p \graycell\epsilon&p \graycell\epsilon&p \graycell\epsilon&p \graycell\epsilon&p \graycell\epsilon&p \graycell\epsilon\\
p \graycell\epsilon&0&p \graycell\epsilon&p \graycell\epsilon&p \graycell\epsilon&p \graycell\epsilon&p \graycell\epsilon&p \graycell\epsilon&p \graycell\epsilon\\
p \graycell\epsilon&p \graycell\epsilon&0&p \graycell\epsilon&p \graycell\epsilon&p \graycell\epsilon&p \graycell\epsilon&p \graycell\epsilon&p \graycell\epsilon\\
p \graycell\epsilon&p \graycell\epsilon&p \graycell\epsilon&0&p \graycell\epsilon&p \graycell\epsilon&p \graycell\epsilon&p \graycell\epsilon&p \graycell\epsilon\\
p \graycell\epsilon&p \graycell\epsilon&p \graycell\epsilon&p \graycell\epsilon&0&p \graycell\epsilon&p \graycell\epsilon&p \graycell\epsilon&p \graycell\epsilon\\
p \graycell\epsilon&p \graycell\epsilon&p \graycell\epsilon&p \graycell\epsilon&p \graycell\epsilon&0&p \graycell\epsilon&p \graycell\epsilon&p \graycell\epsilon\\
p \graycell\epsilon&p \graycell\epsilon&p \graycell\epsilon&p \graycell\epsilon&p \graycell\epsilon&p \graycell\epsilon&0&p \graycell\epsilon&p \graycell\epsilon\\
p \graycell\epsilon&p \graycell\epsilon&p \graycell\epsilon&p \graycell\epsilon&p \graycell\epsilon&p \graycell\epsilon&p \graycell\epsilon&0&p \graycell\epsilon\\
p \graycell\epsilon&p \graycell\epsilon&p \graycell\epsilon&p \graycell\epsilon&p \graycell\epsilon&p \graycell\epsilon&p \graycell\epsilon&p \graycell\epsilon&0\\
\end{array}\right]}_{\textstyle \Exp[Q_\cP^{-1}Q_\cP^c]}
\end{equation*}
\caption{Illustration of individual matrices in Example 1 in Section \ref{sec:uniform}}
\label{fig:uniform2}
\end{figure*}

To derive $\beta, c, \epsilon$ from $\alpha$ we proceed as follows: Let $Q$, $Q_\cP$ and $Q_{\cP^c}$ be given as in Figure~\ref{fig:uniform2}.
The elements of $Q_\cP^{-1}$ can be derived from the individual blocks of $Q_\cP$. $Q_\cP^{-1}$ will have block diagonal structure where the individual blocks satisfy
\[\begin{bmatrix}
1&\alpha&\alpha\\
\alpha&1&\alpha\\
\alpha&\alpha& 1
\end{bmatrix}^{-1}
=
\begin{bmatrix}
c&\beta&\beta\\
\beta&c&\beta\\
\beta&\beta& c
\end{bmatrix}.\]
We denote the diagonal values of each block by $c$ and the off-diagonal elements by $\beta$, where we have
\begin{align*}
\beta &=- \frac 1 {(n_k-2)+\frac 1 \alpha-(n_k-1)\alpha}\\
c&=\frac 1 \alpha \frac{(n_k-2)\alpha +1}{(n_k-2)+\frac 1 \alpha-(n_k-1)\alpha}
\end{align*}
\begin{proof}[Derivation]
Using $Q_\cP Q_\cP^{-1}=I$ we can derive the values of  $\beta$ and $c$ from $\alpha$: We get the two equations $c+(n_k-1)\alpha\beta=1$ for the diagonal elements and $\beta +c\alpha + (n_k-2) \alpha \beta=0$ for the off-diagonal elements. Solving these equations for $c$ and $\beta$ yields the claimed values.
\end{proof}

Further, multiplying $Q_\cP^{-1}$ and $Q_{\cP^c}$ yields a matrix with zero-diagonal blocks and equal non-zero elements as illustrated in Figure~\ref{fig:uniform2} where
\begin{align}
\epsilon& = \frac{1-\alpha}{(n_k-2)+\frac 1 \alpha-(n_k-1)\alpha}= \frac{1-\alpha}{(1-\alpha) n_k+\alpha-2+\frac 1 \alpha}.
\label{eq:epsilon}
\end{align} 
Hence, with $\alpha+\frac 1 \alpha\geq 2$ for $\alpha\leq 1$ we find $\epsilon\leq \frac 1 {n_k}$.

\begin{figure*}[h!]
\includegraphics[width=\columnwidth]{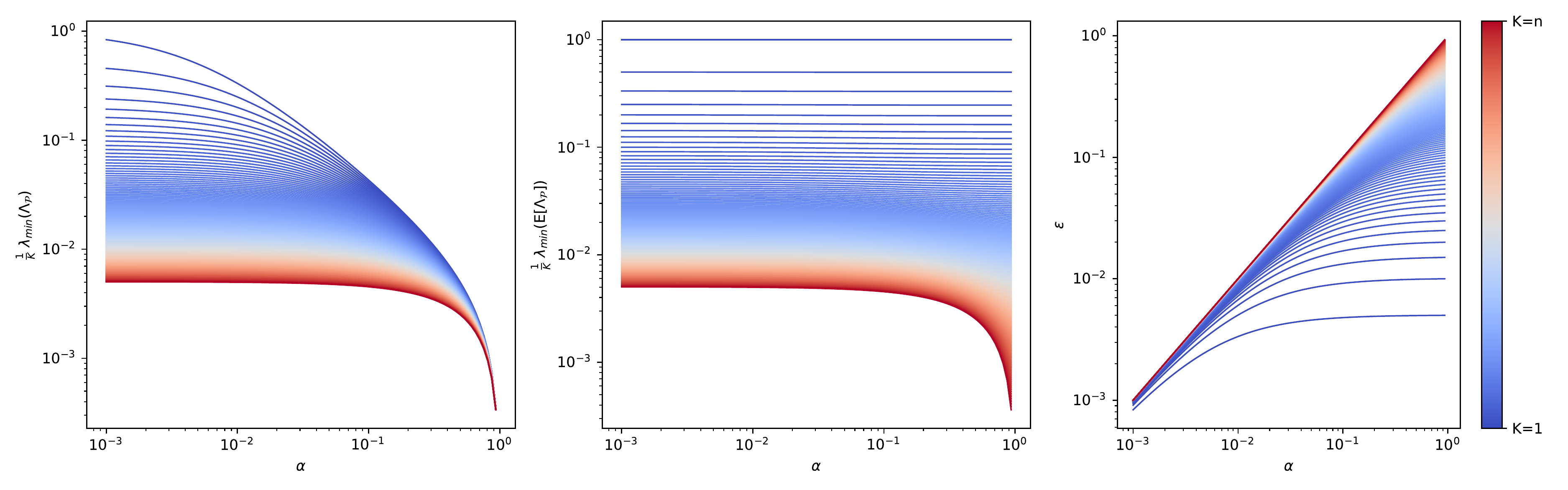}
\caption{Plotting value of $\rho:=\frac 1 K \lambda_{\min}(\Exp[\Lambda_\cP])$  which determines the convergence rate of Algorithm~\ref{alg:algo} (see Theorem~\ref{thm:quad}) . We show $\rho$ for fixed partitioning (left) and for repartitioning (middle), these values are determined by $\epsilon$ (right) as given in \eqref{eq:epsilon}. We use $n=200$, $\alpha\in(0,1)$ and $K\in[1,200]$.  We can make the following two observations: 
1) For small $K$ the rate of repartitioning does not change much with $\alpha$, whereas the rate for static partitioning significantly decreases with increasing $\alpha$. 
2) For large $\alpha$ the rate of static partitioning does not change much with $K$ whereas repartitioning significantly improves for smaller $K$. 
Both findings are verified in the empirical performance of training a ridge regression model as illustrated in Figure~\ref{fig:perfuniform} .}
\label{fig:heatmapeiguniform}
\end{figure*}

The sensitivity of $\epsilon$ and the resulting eigenvalues $\lambda_{\min}(\Lambda_\cP)$ w.r.t. $\alpha$ and $K$ is illustrated in Figure~\ref{fig:heatmapeiguniform}.

\subsection{Separable Data -- Additional Example}
\label{app:separable}

Let us consider  the toy example where $n=4$, $K=2$ and $Q$ has the following separable form:

\begin{figure*}[h!]
\small
\begin{equation*}
\setlength\arraycolsep{3.7pt}
\underbrace{\left[\begin{array}{cccc}
\redcell1&\redcell\alpha&0&0\\
\redcell\alpha&\redcell1&0&0\\
0&0&\redcell1&\redcell\alpha\\
0&0&\redcell\alpha&\redcell1\\
\end{array}\right]}_{\textstyle Q}
\quad
\underbrace{\left[\begin{array}{cccc}
\redcell1&0&0&0\\
0&\redcell1&0&0\\
0&0&\redcell1&0\\
0&0&0&\redcell1\\
\end{array}\right]}_{\textstyle Q_{\cP_1}}
\quad
\underbrace{\left[\begin{array}{cccc}
\redcell1&0&0&0\\
0&\redcell1&0&0\\
0&0&\redcell1&0\\
0&0&0&\redcell1\\
\end{array}\right]}_{\textstyle Q_{\cP_2}}
\quad
\underbrace{\left[\begin{array}{cccc}
\redcell1&\redcell\alpha&0&0\\
\redcell\alpha&\redcell1&0&0\\
0&0&\redcell1&\redcell\alpha\\
0&0&\redcell\alpha&\redcell1\\
\end{array}\right]}_{\textstyle Q_{\cP_3}}
\end{equation*}\\
\begin{equation*}
\setlength\arraycolsep{3.7pt}
\underbrace{\left[\begin{array}{cccc}
\greencell \frac 2 3 +\frac 1 3 c&\greencell\frac 13\beta&0&0\\
\greencell\frac 13\beta&\greencell \frac 2 3 +\frac 1 3 c&0&0\\
0&0&\greencell \frac 2 3 +\frac 1 3 c&\greencell\frac 13\beta\\
0&0&\greencell\frac 13\beta&\greencell \frac 2 3 +\frac 1 3 c\\
\end{array}\right]}_{\textstyle \Exp[Q_{\cP}^{-1}]}
\quad
\underbrace{\left[\begin{array}{cccc}
\greencell1&0&0&0\\
0&\greencell1&0&0\\
0&0&\greencell1&0\\
0&0&0&\greencell1\\
\end{array}\right]}_{\textstyle Q_{\cP_1}^{-1}}
\quad
\underbrace{\left[\begin{array}{cccc}
\greencell1&0&0&0\\
0&\greencell1&0&0\\
0&0&\greencell1&0\\
0&0&0&\greencell1\\
\end{array}\right]}_{\textstyle Q_{\cP_2}^{-1}}
\quad
\underbrace{\left[\begin{array}{cccc}
\greencell c&\greencell\beta&0&0\\
\greencell\beta&\greencell c&0&0\\
0&0&\greencell c&\greencell\beta\\
0&0&\greencell\beta&\greencell c\\
\end{array}\right]}_{\textstyle Q_{\cP_3}^{-1}}
\end{equation*}
\caption{Small example for separable data as discussed  in Section \ref{sec:separable}.}
\label{fig:sep2}
\end{figure*}

Assuming all non-zero off diagonal elements are equal to $\alpha$, there are three possible partitionings which are equally likely in Algorithm~\ref{alg:algo}. The three partitionings are illustrated in Figure~\ref{fig:sep2}. For the two partitionings $\cP_1$ and $\cP_2$ we have 
\[\lambda_{\min}(\Lambda_{\cP_1}) = \lambda_{\min}(\Lambda_{\cP_2}) = 1-\alpha\]
For the third partitioning $\cP_3$ we have 
\[\lambda_{\min}(\Lambda_{\cP_3}) = 1\]
To contrast this with repartitioning, we need to first evaluate $Q_{\cP_3}^{-1}$ where we find $\beta = -\frac 1 {1-\alpha^2}$ and  $c=\frac 1 {1-\alpha^2}$. Hence, we have
\[\lambda_{\min}(\Exp[\Lambda_{\cP}]) =  \frac 1 3 +\frac 2 3 (1-\alpha)\]
As a consequence, repartitioning will perform better than static partitioning in two out of three cases for this separable toy example. Hence if we are not able to recover the optimal partitioning $\cP^\star$ we would be better off to use Algorithm~\ref{alg:algo} with repartitioning.
\newpage

\section{Additional Experiments}
\label{app:add}

\textbf{Sensitivity of $\lambda_{\min}(\Lambda_\cP)$ and $\lambda_{\min}(\Exp[\Lambda_\cP])$ computed in Section \ref{sec:uniform} w.r.t $K$ and $\alpha$\\}

\textbf{Effect of using  line search vs fixed step size\\}

\begin{figure*}[h!]
\centering
\subfigure[separable ($\alpha=0.6, K=5$)]{\includegraphics[width=0.32\columnwidth]{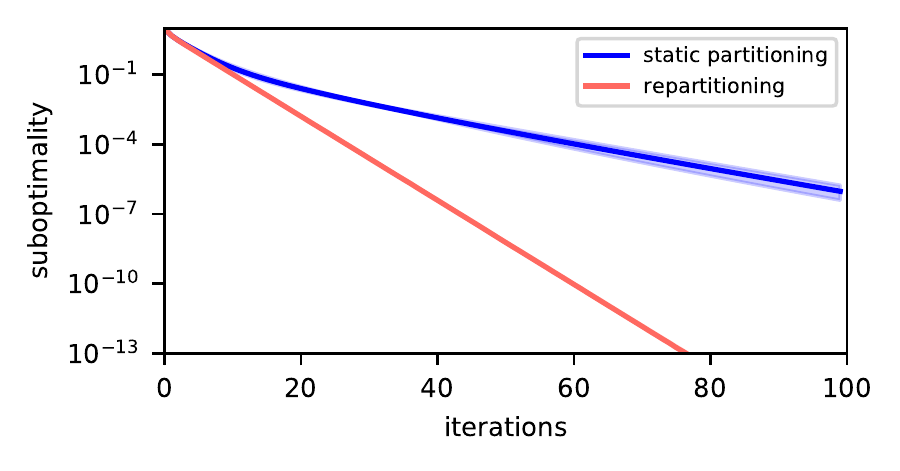}}
\subfigure[uniform data ($\alpha=0.4, K=5$)]{\includegraphics[width=0.32\columnwidth]{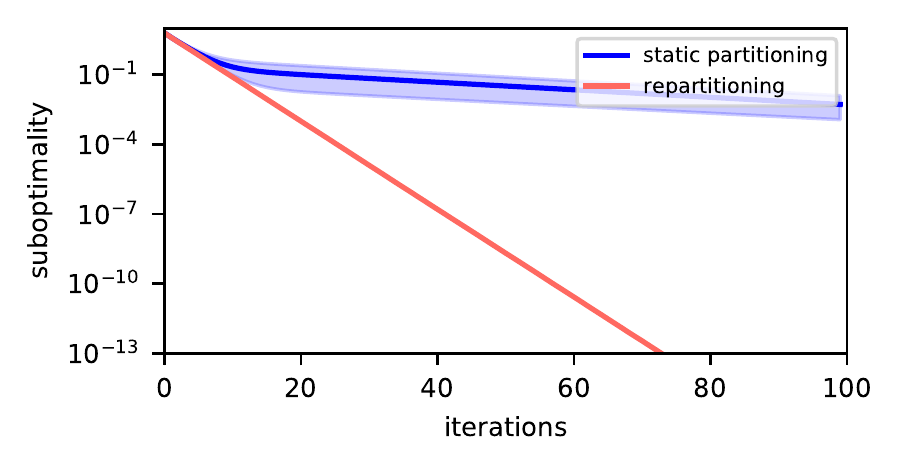}}
\subfigure[random data ($K=8$)]{\includegraphics[width=0.32\columnwidth]{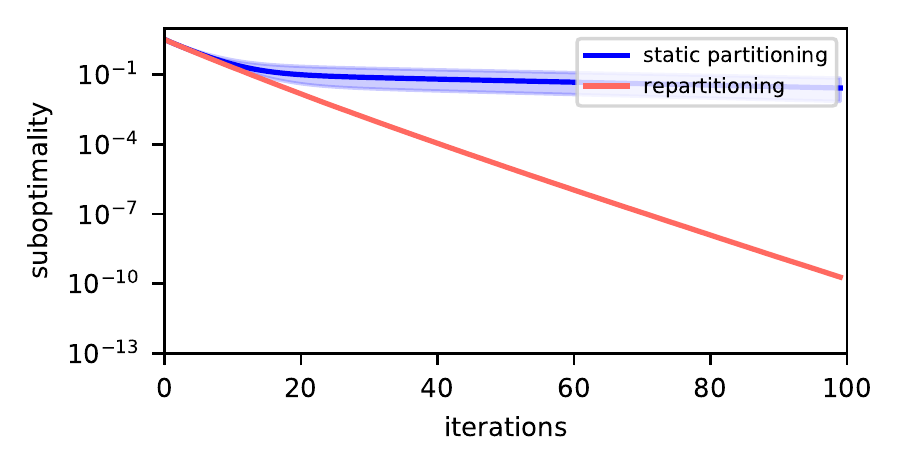}}
\subfigure[separable ($\alpha=0.6, K=5$)]{\includegraphics[width=0.32\columnwidth]{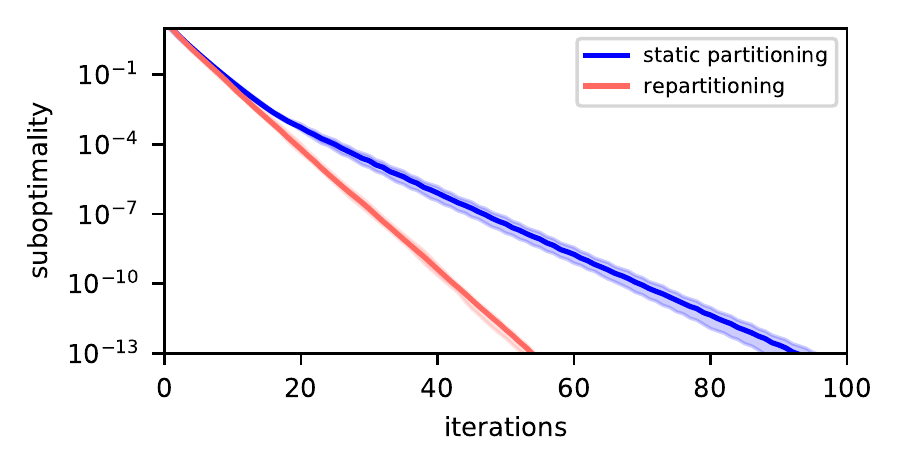}}
\subfigure[uniform data ($\alpha=0.4$)]{\includegraphics[width=0.32\columnwidth]{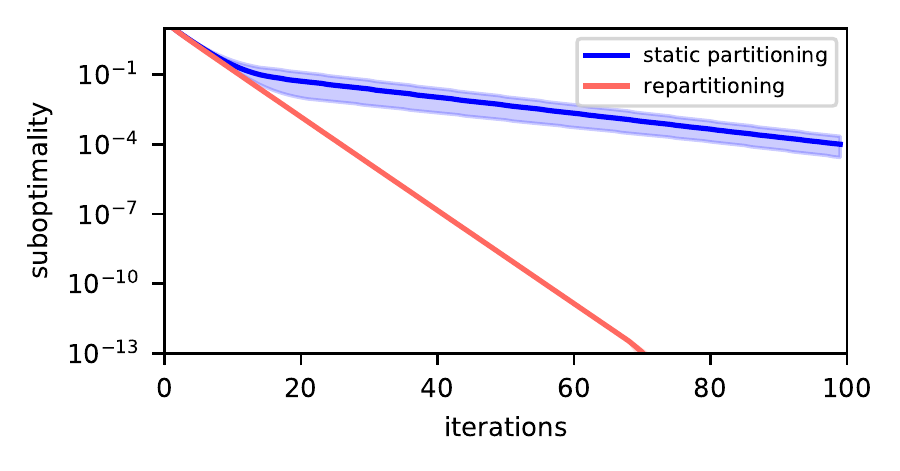}}
\subfigure[random data ($K=8$)]{\includegraphics[width=0.32\columnwidth]{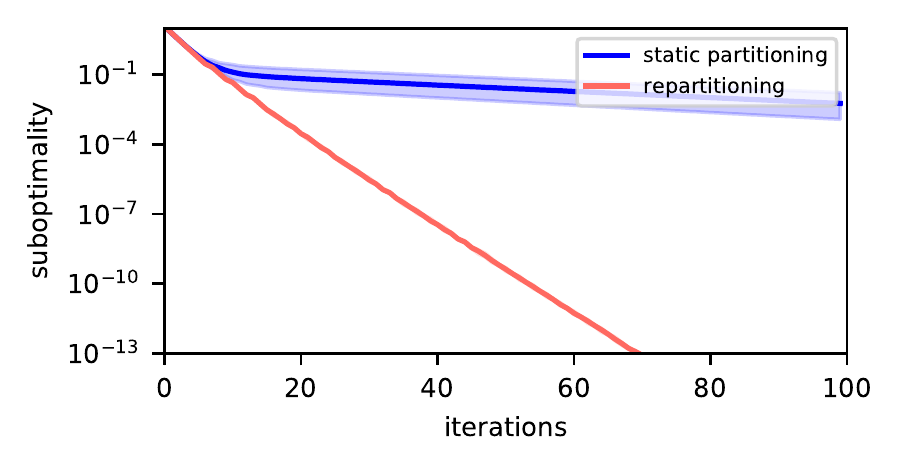}}
\caption{Investigating the effect of line search on the performance of Algorithm~\ref{alg:algo} with static and dynamic partitioning for three different datasets: separable data as discussed in Section~\ref{sec:separable}, data with uniform correlation as discussed in Section~\ref{sec:uniform} and random data, where each entry of $A$ is drawn from a random normal distribution. Confidence intervals show min-max-intervals over 10 repetitions. The top line shows performance with fixed step size $\eta = \frac 1 K$ and the bottom line shows performance with line-search. We see that the relative behavior of static and dynamic partitioning is preserved across all datasets, justifying the study of a fixed step size in the main part of the paper.}
\label{fig:ls}
\end{figure*}

\newpage
\textbf{Convergence of ADN \citep{duenner2018adn} with and without repartitioning as a function of the number of processes\\}

\begin{figure*}[h!]
\centering
\subfigure{\includegraphics[width=0.245\columnwidth]{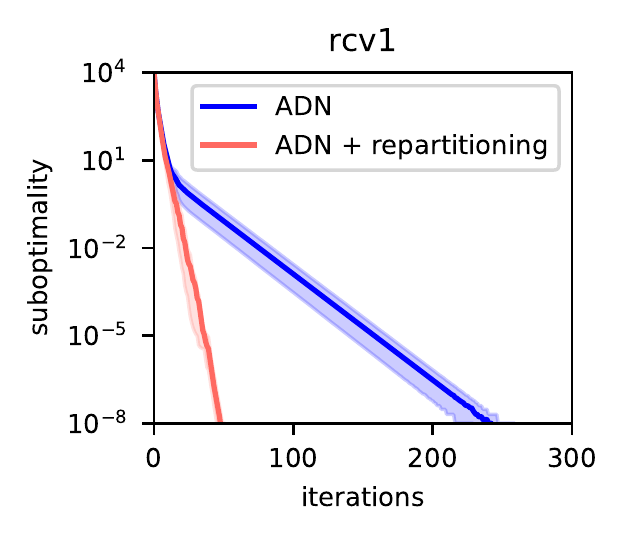}}
\subfigure{\includegraphics[width=0.245\columnwidth]{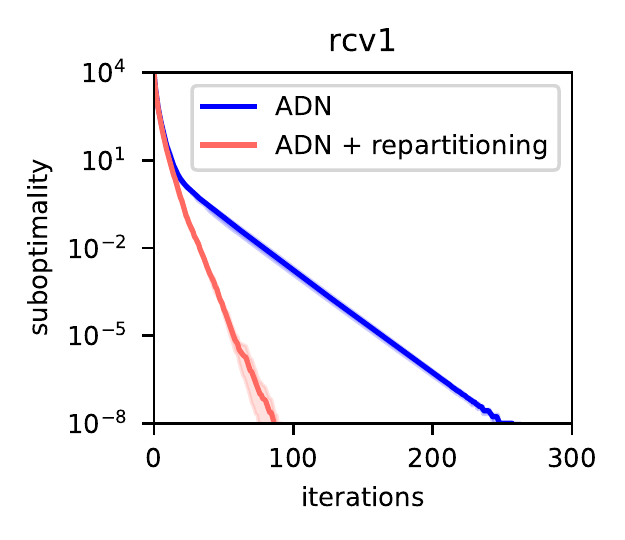}}
\subfigure{\includegraphics[width=0.245\columnwidth]{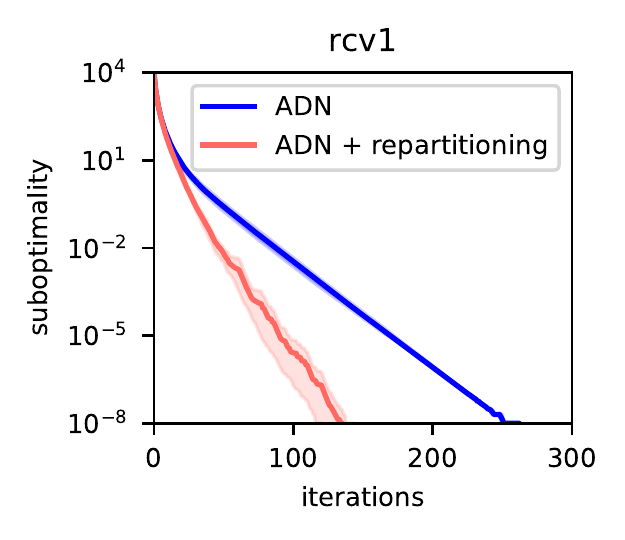}}
\subfigure{\includegraphics[width=0.245\columnwidth]{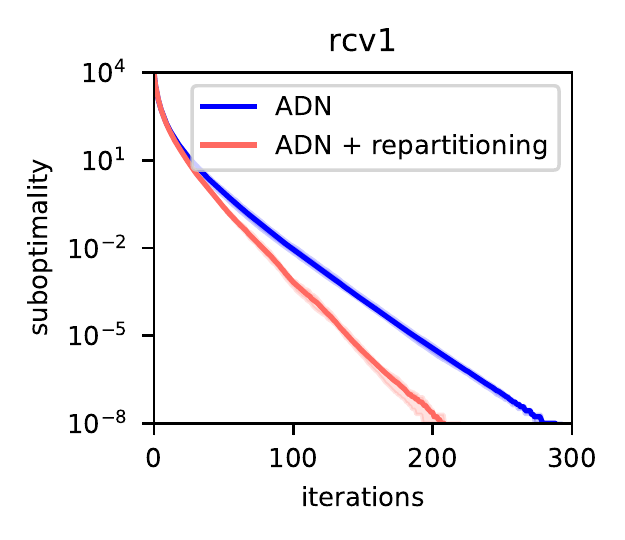}}\\
\subfigure{\includegraphics[width=0.245\columnwidth]{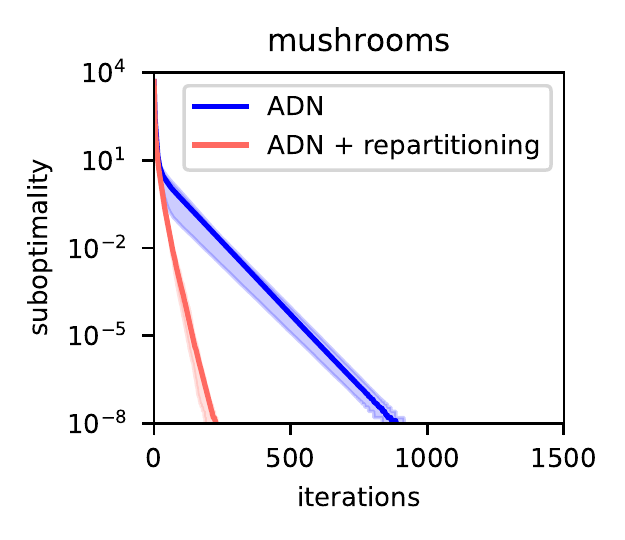}}
\subfigure{\includegraphics[width=0.245\columnwidth]{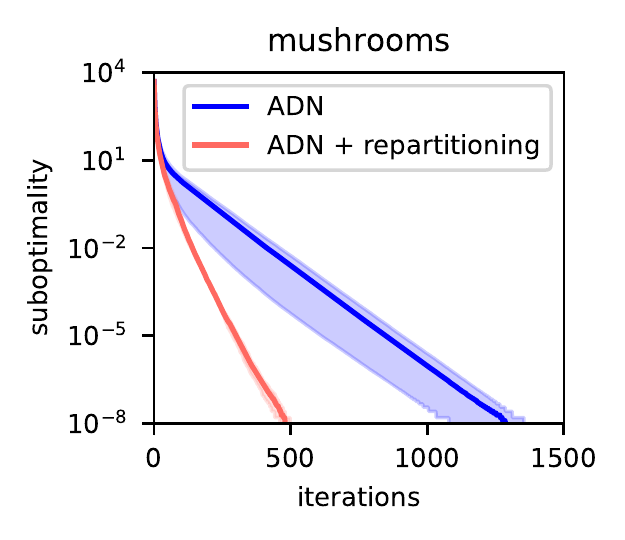}}
\subfigure{\includegraphics[width=0.245\columnwidth]{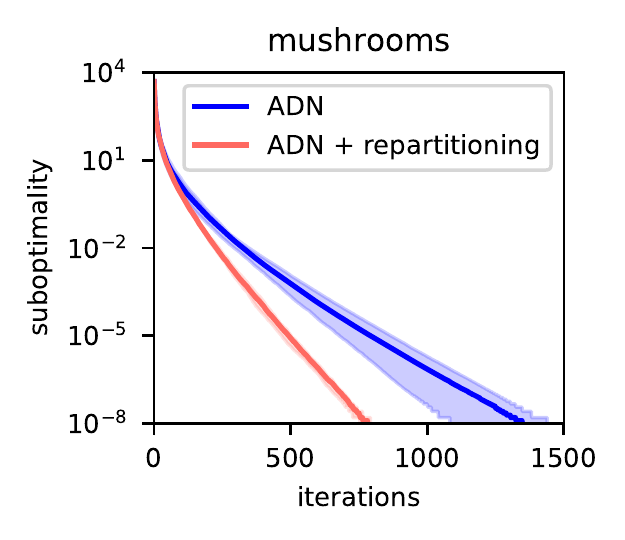}}
\subfigure{\includegraphics[width=0.245\columnwidth]{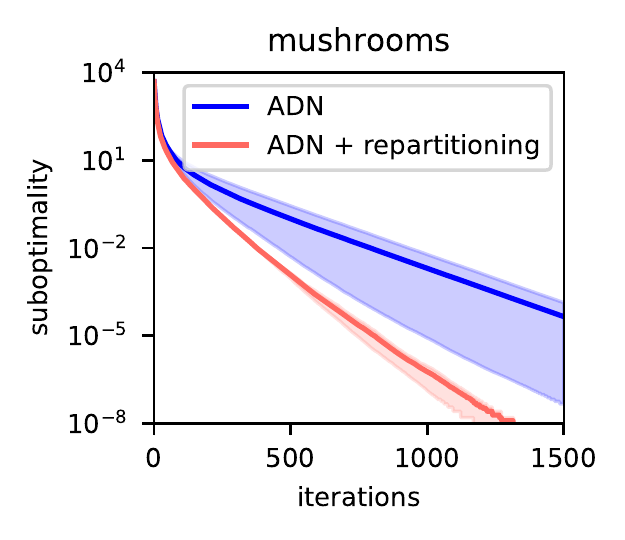}}\\
\subfigure{\includegraphics[width=0.245\columnwidth]{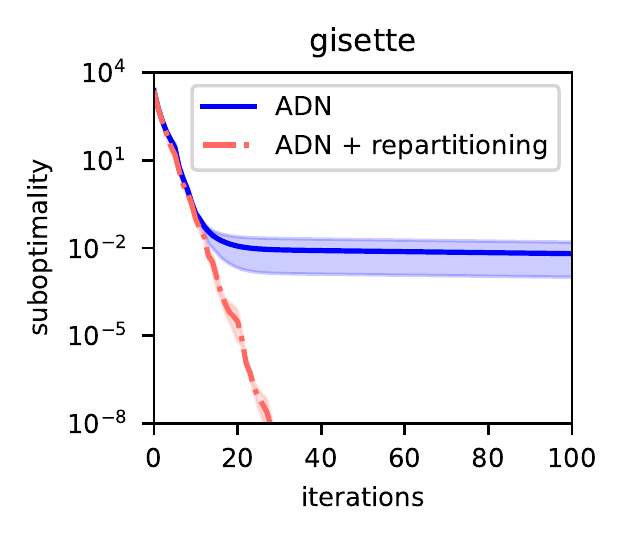}}
\subfigure{\includegraphics[width=0.245\columnwidth]{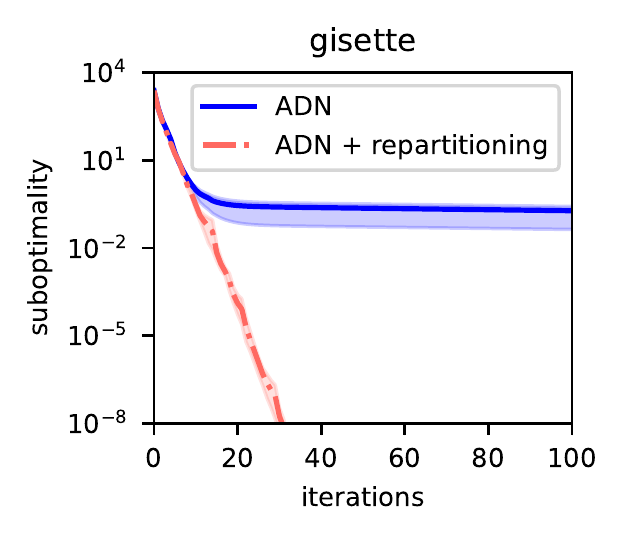}}
\subfigure{\includegraphics[width=0.245\columnwidth]{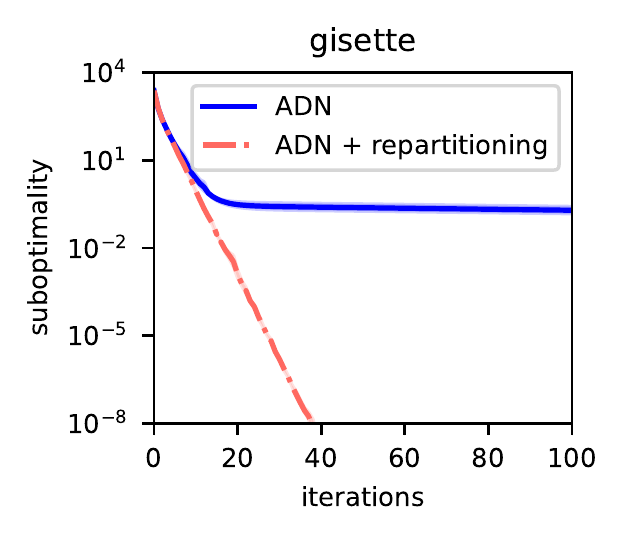}}
\subfigure{\includegraphics[width=0.245\columnwidth]{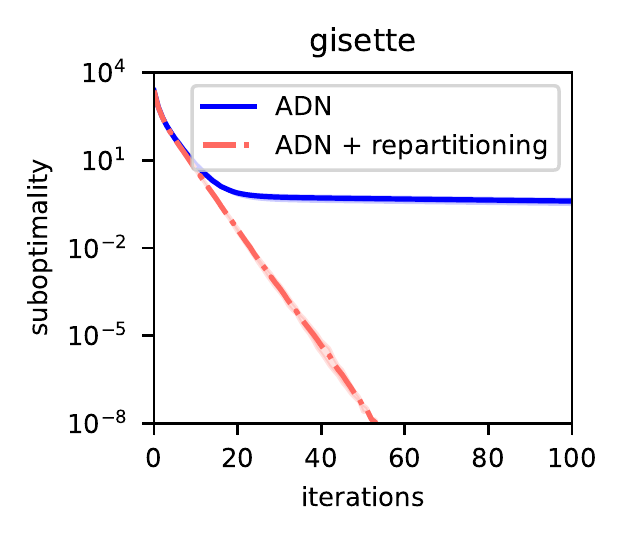}}\\
 \addtocounter{subfigure}{-12}
\subfigure[$K=2$]{\includegraphics[width=0.245\columnwidth]{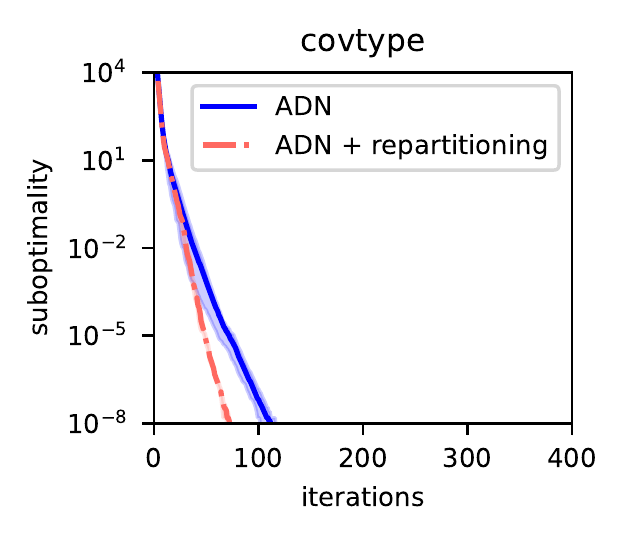}}
\subfigure[$K=4$]{\includegraphics[width=0.245\columnwidth]{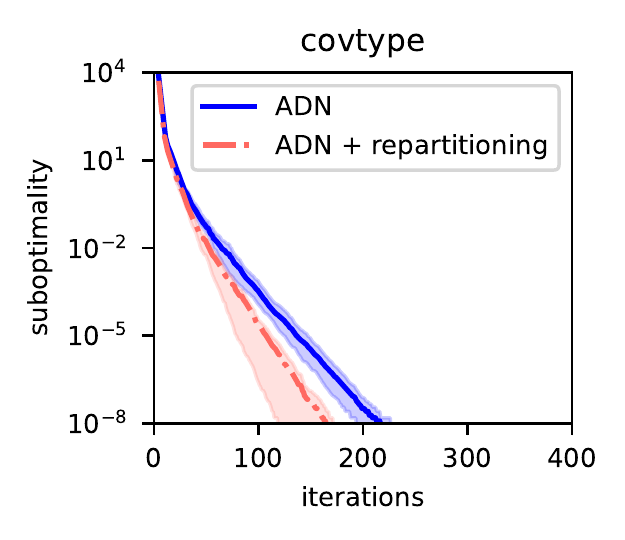}}
\subfigure[$K=8$]{\includegraphics[width=0.245\columnwidth]{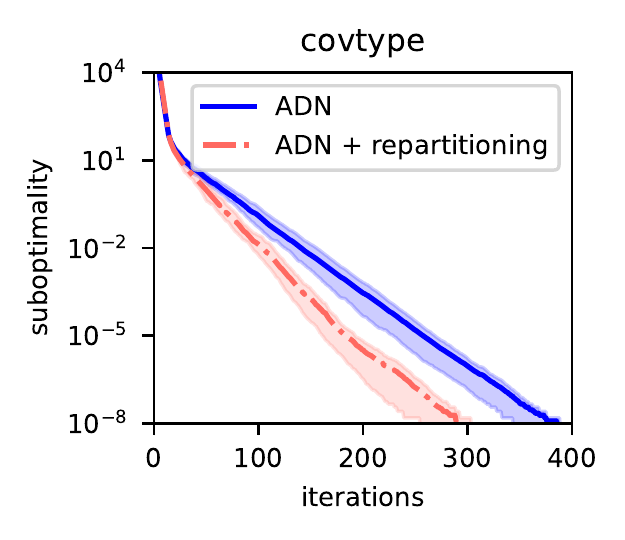}}
\subfigure[$K=16$]{\includegraphics[width=0.245\columnwidth]{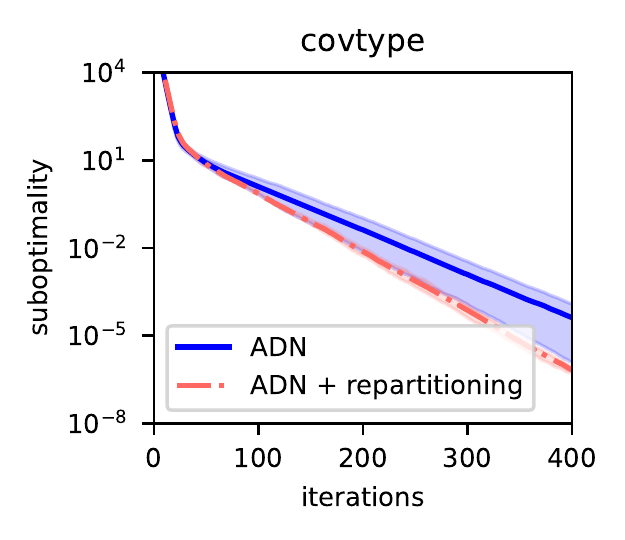}}
\caption{Convergence of ADN \citep{duenner2018adn} with and without repartitioning for different datasets and values of $K$. Confidence intervals show min-max-intervals over 10 repetitions. We see that repartitioning achieves a significant gain over static partitioning across all datasets and for different number of $K$. The performance of static partitioning can be sensitive to the quality of the partitioning. This is the case for datasets with highly non-uniform features, such as covtype and mushroom. In the covtype dataset the features are a mix of $11$ real valued and $43$ categorical features. In the mushroom dataset, the sparsity of the features varies a lot; over $50\%$ of the features are more than $80\%$ sparse and the other half of the features covers all the spectrum up to fully dense features. }
\label{fig:ADNapp}
\end{figure*}

\newpage
\textbf{Theoretical and empirical convergence on random data\\}

\begin{figure*}[h!]
\centering
\subfigure[$\alpha=0.01$]{\includegraphics[width=0.32\columnwidth]{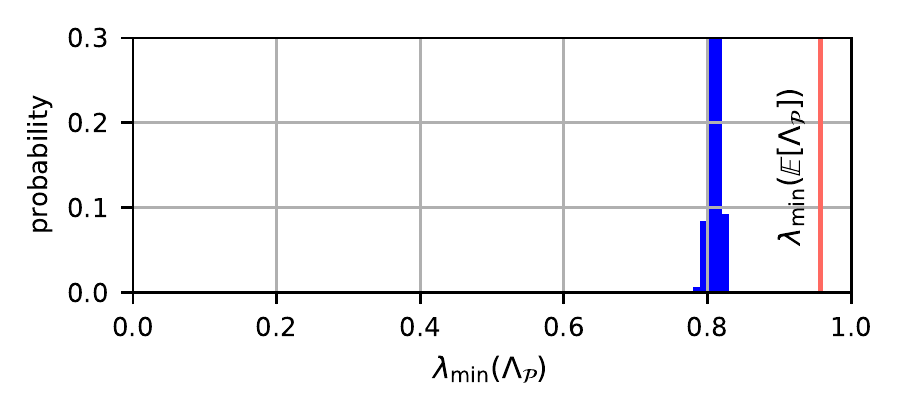}}
\subfigure[$\alpha=0.05$]{\includegraphics[width=0.32\columnwidth]{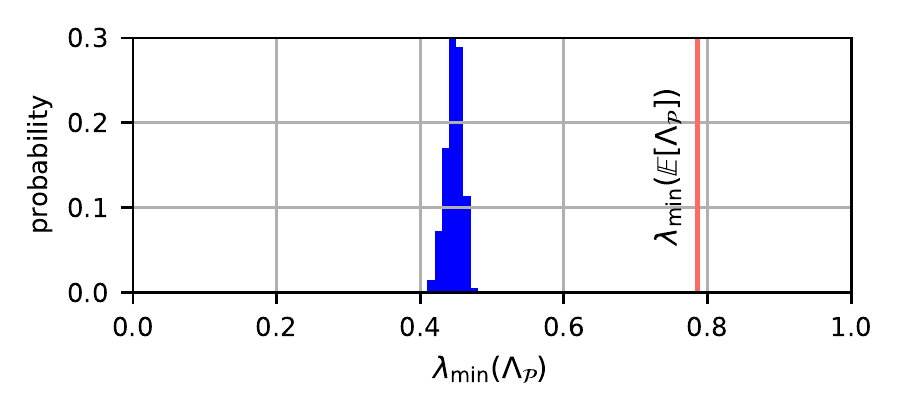}}
\subfigure[$\alpha = 0.1$]{\includegraphics[width=0.32\columnwidth]{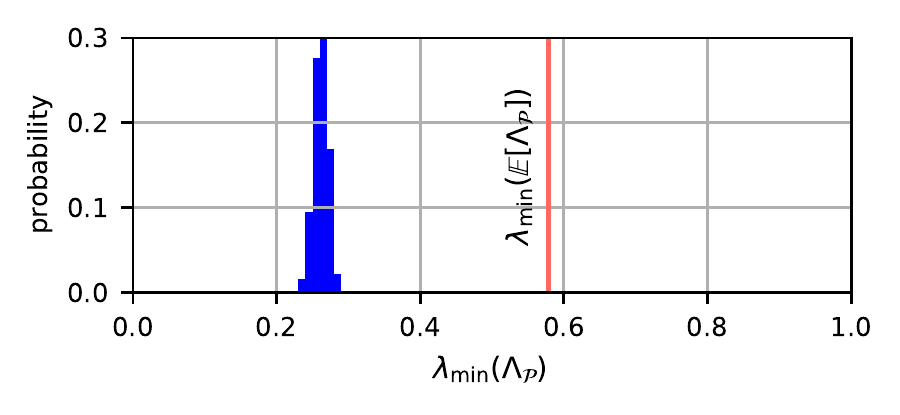}}
\subfigure[$\alpha = 0.01$]{\includegraphics[width=0.32\columnwidth]{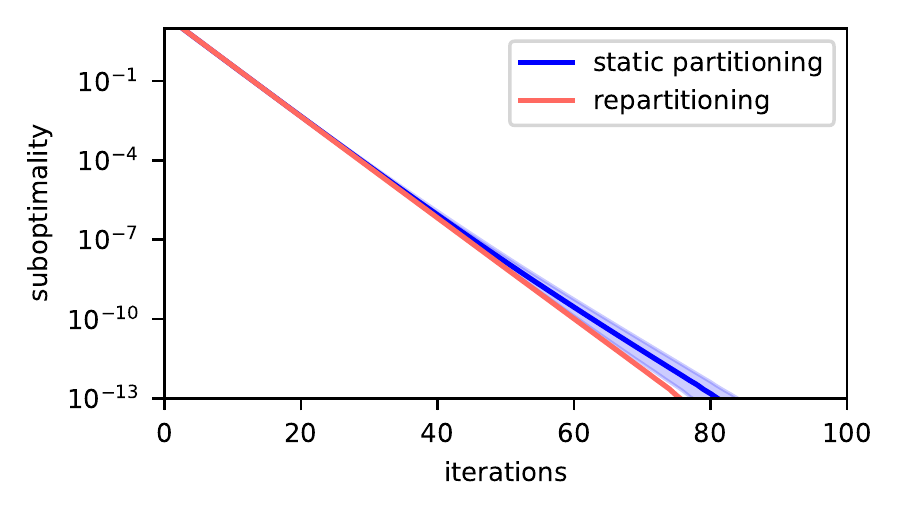}}
\subfigure[$\alpha=0.05$]{\includegraphics[width=0.32\columnwidth]{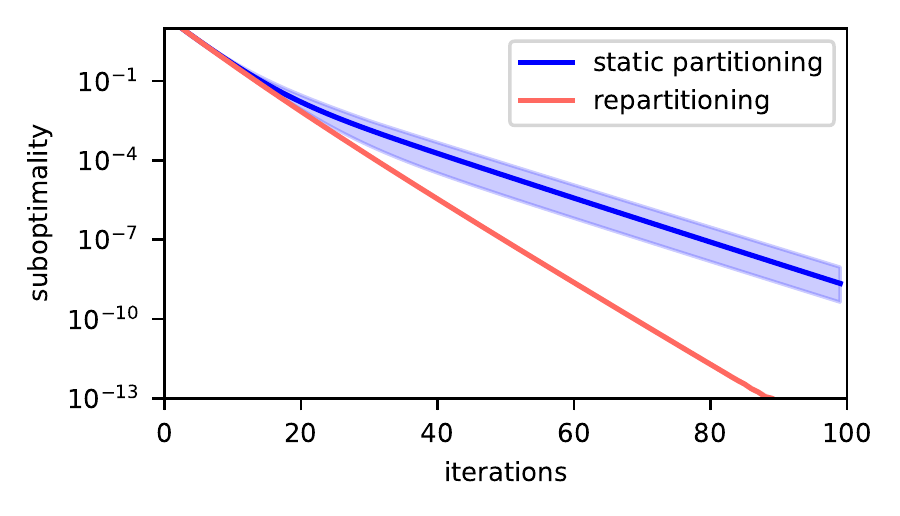}}
\subfigure[$\alpha = 0.1$]{\includegraphics[width=0.32\columnwidth]{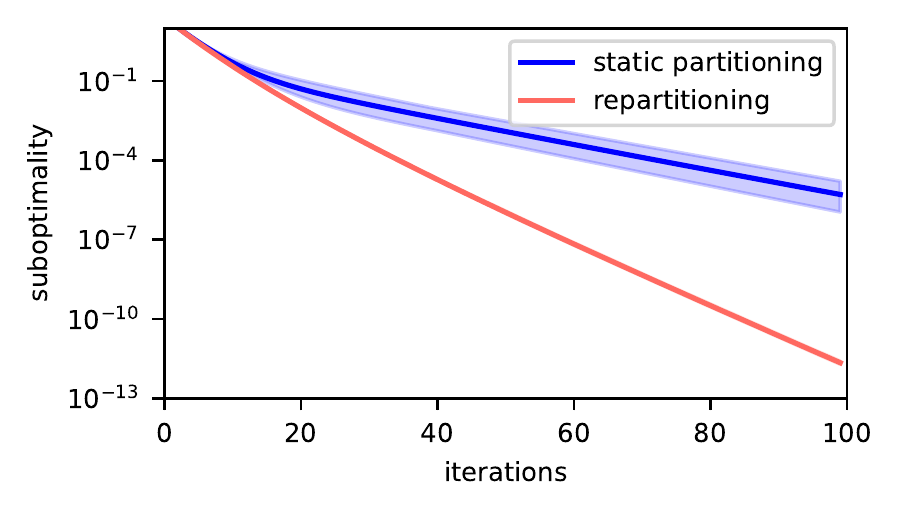}}
\caption{Theoretical and empirical convergence of Algorithm~\ref{alg:algo} with and without repartitioning for random data with varying correlation strength $\alpha$. Data was generated by sampling the elements of $A^\top A\sim \mathcal N(\alpha,\frac \alpha 2)$ in a symmetric fashion. Top figures show the distribution of $\lambda_{\min}(\Lambda_\cP)$  (determining the rate of Algorithm~\ref{alg:algo} for static partitioning) across 1000 random partitionings $\cP$ in comparison to $\lambda_{\min}(\Exp[\Lambda_\cP])$ (determining the rate of Algorithm~\ref{alg:algo} for repartitioning). The figures on the bottom show the corresponding empirical convergence for training a ridge regression classifier. We see that the ratio between the eigenvalues explains the convergence difference observed empirically. }
\label{fig:hist_app}
\end{figure*}

\textbf{Effect of regularization on theoretical performance gain of repartitioning through $\lambda_{\min}(\Exp[\Lambda_\cP])$\\}
\begin{figure*}[h!]
\centering
\subfigure{\label{fig:violinmushroom}\includegraphics[width=0.49\columnwidth]{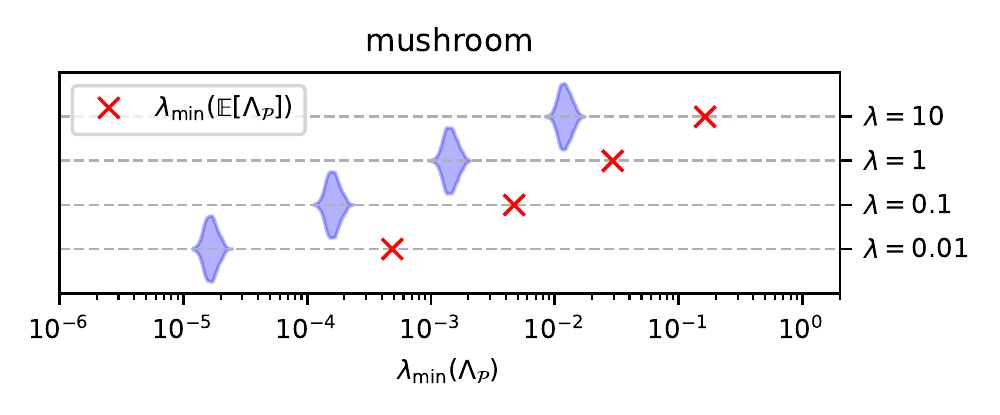}}
\subfigure{\label{fig:violinsyn}\includegraphics[width=0.49\columnwidth]{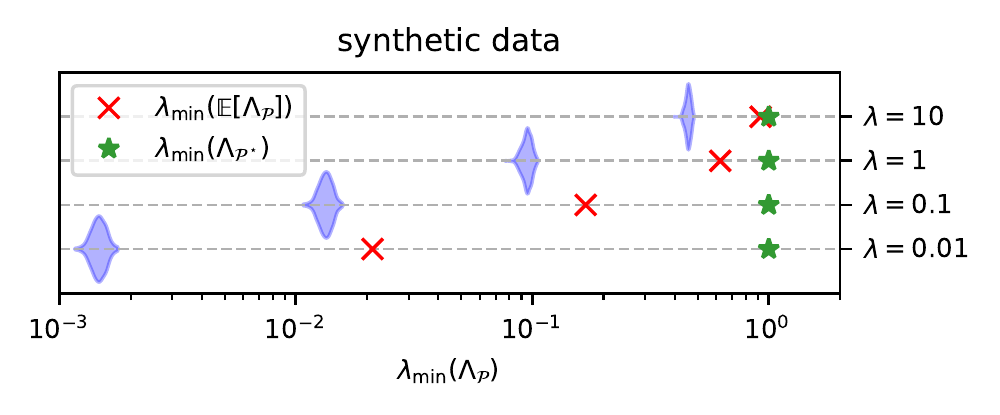}}
\caption{ Effect of regularization on the distribution of $\lambda_{\min}(\Lambda_\cP)$ across 1000 random partitions on different datasets with $K=5$; mushroom data (left) and synthetic data with $A^\top A\sim \mathcal N(\alpha,\frac \alpha 2)$ for $\alpha=0.5$ (right). We consider $Q = A^\top A + \lambda I$ such as in linear regression for varying $\lambda$. We compare $\lambda_{\min}(\Lambda_\cP)$ that determines the rate of static partitioning to $\lambda_{\min}(\Exp[\Lambda_\cP])$ that governs the rate of dynamic partitioning and, if known, $\lambda_{\min}(\Lambda_{\cP^\star}$) that determines the convergence of the best static  partitioning. Note that for synthetic data, adding regularization has the same effect than decreasing $\alpha$.}
\label{fig:violinapp}
\end{figure*}

\newpage

\section{Discussion on Model Assumption}

In Section \ref{sec:conv} we consider  the following assumption on the auxiliary model:
\begin{equation}
\label{ass}
f(\xvt + \Dxv)\leq \xi \tilde f_\xvt(\Dxv, Q_t) + (1-\xi) f(\xvt)
\end{equation}
Recall that the model is defined as 
\begin{align}
\tilde f_\xvt(\Dxv;Q_t):=f(\xvt) + \nabla f(\xvt)^\top \Dxv + \frac 1 {2} \Delta \xv^\top Q_t \Delta\xv.\notag
\end{align}
In the following we will outline how some popular distributed methods fit into this framework:

\subsection{CoCoA }
The auxiliary model $\tilde f_\xv$ in CoCoA \citep{Smith:2016wp} is designed for GLMs and defines 
\[Q_t=\gamma_\ell A^\top A\]
 where $\gamma_\ell$ denotes the smoothness parameter of the loss function. Crucial for the algorithm is that this choice guarantees that the model forms a global upper bound on the function $f$. As a consequence it satisfies Assumption \eqref{ass} for $\xi=1$.

\subsection{Line Search}

Methods such as \citep{lee2017distributed} use the true Hessian $\nabla^2f(\xv)$ and deploy a line search strategy to rescale the update by $\beta_t$ and guarantee sufficient function decrease. The rescaling of the update can be absorbed into $Q_t$ which then becomes 
\[Q_t:=\frac 1 \beta_t \nabla^2f(\xv).\]
 In that way we offload the concerns about convergence to the choice of the auxiliary model which is outside the scope of our study. The backtracking lines search control parameter $\alpha$ then corresponds exactly to $\xi$ and $\beta_t$ is equal to the corresponding step size satisfying the required stopping criteria.

To see this, consider the stopping criteria of linesearch:
\[f(\xv+\Dxv)\leq f(\xv) +\alpha \nabla f(\xv)^\top \Dxv\]
and hence for all $Q$ PSD and $\alpha>0$ it holds that
\begin{align} f(\xv+\Dxv)&\leq f(\xv) +\alpha \nabla f(\xv)^\top \Dxv + \alpha \frac 1 2 \Dxv^\top Q_t \Dxv\notag\\
&= (1-\alpha) f(\xv) +\alpha \left[f(\xv) + \nabla f(\xv)^\top \Dxv +\frac 1 2 \Dxv^\top Q_t \Dxv\right]\notag\\
&= (1-\alpha) f(\xv) +\alpha \tilde f_\xv(\Dxv , Q_t) 
\end{align}

\subsection{Trust Region}
A trust region approach such as ADN proposed in \citep{duenner2018adn} is not fully covered by our setting. The challenge is that the trust region approach guarantees that the model decrease is close to the function decrease, it does however not guarantee monotonic improvement. To be more precise, TR acts directly on the update computed by the diagonalized model $\tilde f_\xv (\cdot,Q_\cPt)$ and adjusts $Q_t$ accordinaly. 
Theorem 4.5 in~\citep{convex_optimization_nocedal} shows that for all $t$ sufficiently large, there exists a finite constant $0<c<1$ such that
\begin{align}
| \rho - 1 | &= \left| \frac{f(\xv_t)-f(\xv_t+\Dxv) - (f(\xv_t)-\tilde f_\xv (\cdot,Q_\cPt))}{f(\xv_t)-\tilde f_\xv (\cdot,Q_\cPt)} \right| \\
&= \left | \frac{\tilde f_\xv (\cdot,Q_\cPt)-f(\xv_t+\Dxv)}{f(\xv_t)-\tilde f_\xv (\Dxv,Q_\cPt)} \right| < c.
\end{align}
Therefore,
\begin{align}
f(\xv_t+\Dxv) - \tilde f_\xv (\cdot,Q_\cPt) < c (f(\xv_t)-\tilde f_\xv (\cdot,Q_\cPt)) \\
\implies f(\xv_t+\Dxv) \leq (1-c) \tilde f_\xv (\Dxv,Q_\cPt) + c f(\xv_t)
\end{align}
Our assumption \eqref{ass}, however requires the same bound to hold for  $\tilde f_\xv(\Dxv, Q_t)$ instead of $\tilde f_\xv (\cdot,Q_\cPt)$.
The analysis would hence need to be extended and performed in the trust-region style \citep{nesterov2006cubic} which we do not expect to pose major technical difficulties, nor change the effectiveness of repartitioning significantly. 
This has been confirmed by the experimental results on ADN in Figure \ref{fig:ADNapp}.

\end{document}